\documentclass[preprint]{vgtc}               

\ifpdf
  \pdfoutput=1\relax                   
  \usepackage{graphicx}                
  \DeclareGraphicsExtensions{.pdf,.png,.jpg,.jpeg} 
\else
  \usepackage{graphicx}                
  \DeclareGraphicsExtensions{.eps}     
\fi
\graphicspath{{figures/}{pictures/}{images/}{./}} 

\usepackage{microtype}                 
\PassOptionsToPackage{warn}{textcomp}  
\usepackage{textcomp}                  
\DeclareMathAlphabet{\mathcal}{OMS}{cmsy}{m}{n} 
\usepackage{times}                     
         
\usepackage{cite}                      
\usepackage{tabu}                      
\usepackage{booktabs}                  
\usepackage{float}
\usepackage{longtable,booktabs}
\usepackage[ruled,vlined]{algorithm2e}
\usepackage{multirow}
\usepackage{amsmath}
\usepackage{amsfonts}
\usepackage{mathtools}
\usepackage{xcolor}

\definecolor{Gray}{gray}{0.9}
\definecolor{demphcolor}{RGB}{144,144,144}
\definecolor{backblue}{RGB}{178,231,250}
\definecolor{backblue1}{RGB}{186,216,242}
\definecolor{backred}{RGB}{255,178,178}
\newcommand{\true}[1]{{\setlength{\fboxsep}{0pt}\colorbox{backblue}{#1}}}
\newcommand{\false}[1]{{\setlength{\fboxsep}{0pt}\colorbox{backred}{#1}}}
\onlineid{1036}
\vgtccategory{Research}
\vgtcinsertpkg

\renewcommand{\vec}[1]{\boldsymbol{#1}}

\newcommand{\x}{\vec{x}}

\newcommand{\I}{\vec{I}}
\newcommand{\Imask}{\vec{I}_\text{m}}

\def\eg{\emph{e.g}.} 
\def\ie{\emph{i.e}.} 
 
\def\etc{\emph{etc}.} 
 
\def\etal{\emph{et al}.}

\makeatletter
\newcommand{\thickhline}{%
	\noalign {\ifnum 0=`}\fi \hrule height 1pt
	\futurelet \reserved@a \@xhline
}
\makeatother

\usepackage{tablefootnote} 
\usepackage{orcidlink}

\makeatletter
\newcommand{\savedauthor}{}
\newcommand{\savedaffiliation}{}
\AtBeginDocument{
    \let\savedauthor\@author
    \let\savedaffiliation\vgtc@affiliation
}

\def\@supptitle{}
\newcommand{\supptitle}[1]{\def\@supptitle{#1}} 

\newcommand{\maketitlesupplementary}{%
  \clearpage 
  \onecolumn 
  \setcounter{footnote}{0} 
  \renewcommand{\thefootnote}{\fnsymbol{footnote}} 
  
  \begin{center}
    {\sffamily\LARGE\bfseries\vgtc@sectionfont \@supptitle \par}
    \vspace{0.5em}
    {\mdseries\large\itshape \textemdash\quad Supplementary Materials\quad\textemdash}\par
    \vspace{1.5\baselineskip}
    {\large\sffamily\vgtc@sectionfont 
      \let\and\vgtc@origand  
      \let\thanks\tablefootnote 
      \begin{tabular}[t]{c}
        \savedauthor
      \end{tabular}\par
    }
    \ifx\savedaffiliation\vgtc@empty \else
      \par\vspace{1\baselineskip}
      {\savedaffiliation\par}
    \fi
  \end{center}
  \vspace{\titlespace}
  \thispagestyle{empty} 
}
\makeatother

\title{ProCap: Projection-Aware Captioning for Spatial Augmented Reality}
\supptitle{ProCap: Projection-Aware Captioning for Spatial Augmented Reality}

\author{
    Zimo Cao\orcidlink{0009-0007-4210-2397}\thanks{e-mail: \{caozimo, swudyc714\}@email.swu.edu.cn. Equal contribution.}\\ %
	\scriptsize Southwest University %
    \and Yuchen Deng\orcidlink{0009-0006-3893-955X}\protect\footnotemark[1]\\ %
	\scriptsize Southwest University %
	\and Haibin Ling\orcidlink{0000-0003-4094-8413}\thanks{e-mail: linghaibin@westlake.edu.cn}\\ %
	\scriptsize Westlake University %
	\and Bingyao Huang\orcidlink{0000-0002-8647-5730}\thanks{e-mail: bhuang@swu.edu.cn. Corresponding author.}\\ %
	\scriptsize Southwest University }
\teaser{
  \centering
  \includegraphics[width=1.0\linewidth]{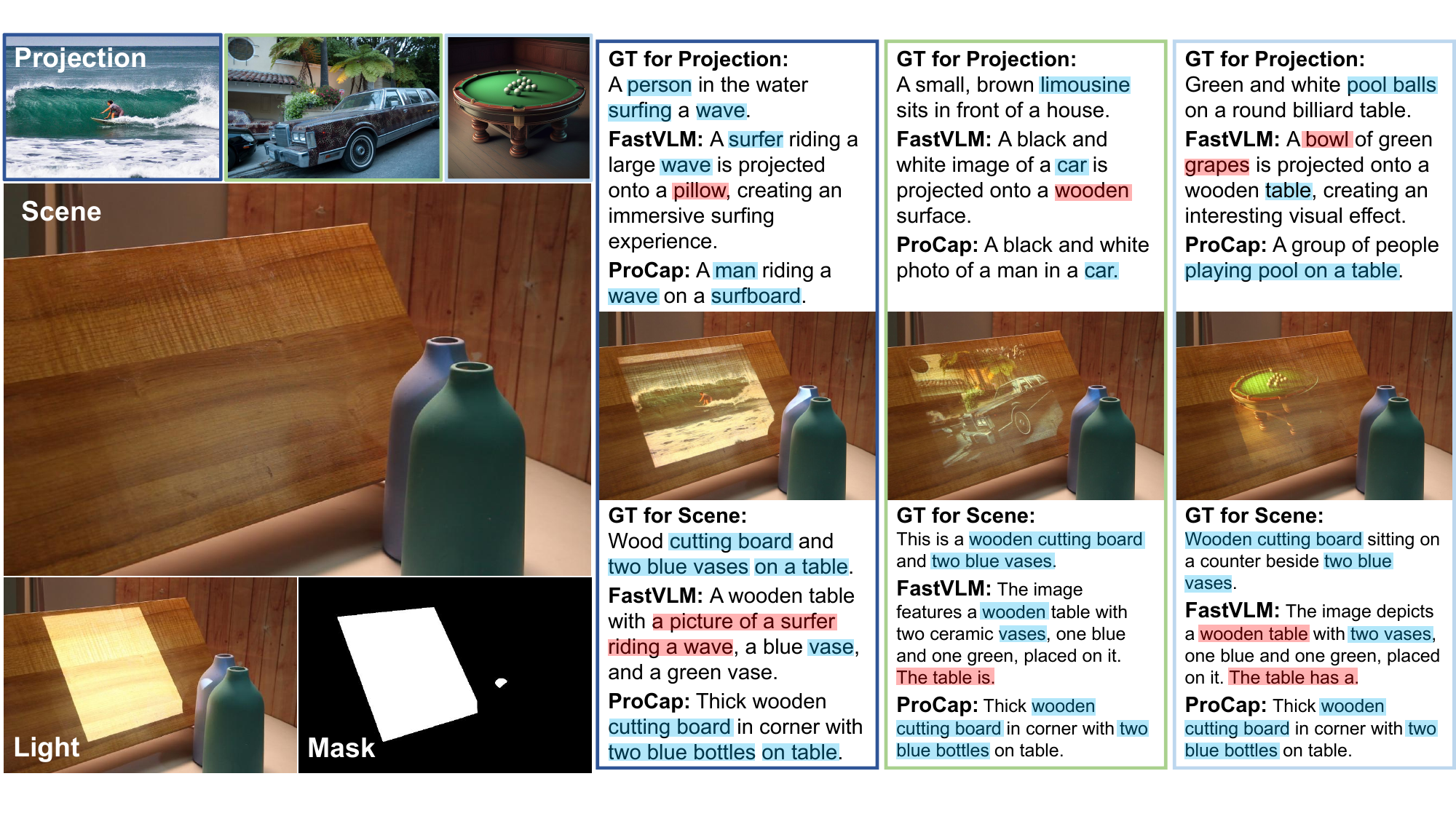}
    \caption{ProCap framework for decoupled SAR scene captioning. Standard VLMs (\eg, FastVLM) often fail to distinguish between \textbf{physical scene} and \textbf{projected content} in SAR, and generate hallucinated captions. By leveraging an automatic segmentation module and region-aware retrieval, ProCap generates distinct, high-fidelity captions for both components (right). We highlight errors in \false{red} and correct descriptions in \true{blue}. Note that the model outputs are truncated by the max tokens limit.}
  \label{fig:teaser}
}
\abstract{Spatial augmented reality (SAR) directly projects digital content onto physical scenes using projectors, creating immersive experience without head-mounted displays. However, for SAR to support intelligent interaction, such as reasoning about the scene or answering user queries, it must semantically distinguish between the physical scene and the projected content. Standard Vision Language Models (VLMs) struggle with this virtual-physical ambiguity, often confusing the two contexts. To address this issue, we introduce ProCap, a novel framework that explicitly decouples projected content from physical scenes. ProCap employs a two-stage pipeline: first it visually isolates virtual and physical layers via automated segmentation; then it uses region-aware retrieval to avoid  ambiguous semantic context due to projection distortion. To support this, we present RGBP (RGB + Projections), the first large-scale SAR semantic benchmark dataset, featuring 65 diverse physical scenes and over 180,000 projections with dense, decoupled annotations. Finally, we establish a dual-captioning evaluation protocol using task-specific tokens to assess physical scene and projection descriptions independently. Our experiments show that ProCap provides a robust semantic foundation for future SAR research. 
The source code, pre-trained models and the RGBP dataset are available on the project page: \url{https://ZimoCao.github.io/ProCap/}.
}

\begin{document}
\maketitle

\section{Introduction}\label{sec:introduction}
Spatial augmented reality (SAR)~\cite{Bimber2005sar, DaisukeIWAI2024pjab.100.012} superimposes digital content directly onto physical scenes using projectors. By modifying local textures and colors with light, SAR enhances the visual properties of physical objects for domains ranging from design and manufacturing~\cite{bermano2017makeuplamps, kaminokado2019augment, Asahina2021realistic, huang2021flexible, dong2023Calibration} to exhibition~\cite{Kageyama2022deblur, grundhofer2015robust,iwai2006limpiddesk, Kagami2019AnimatedStickies, erel2023neuralprojectionmapping, erel2024casperdpm, Deng2025LAPIG}. 
Despite their visual appeal, most existing SAR systems rely on high-fidelity pre-scanned geometries and content specifically authored for fixed scenes. To transition from these rigid displays to autonomous, instruction-aware agents, SAR must achieve a deep semantic understanding of arbitrary scenes and project content dynamically based on user intent and internal reasoning. 

Vision Language Models (VLMs)~\cite{radford2021clip, li2022blip, blip2Qformer, llama3modelcard, fastvlm2025, bai2025qwen25vltechnicalreport, bai2025qwen3vltechnicalreport, zhu2025internvl3exploringadvancedtraining} present a promising foundation for this evolution, enabling multimodal reasoning~\cite{bai2025qwen25vltechnicalreport, bai2025qwen3vltechnicalreport} and natural language interaction within physical spaces. However, as illustrated in~\cref{fig:teaser}, directly applying off-the-shelf VLMs to SAR environments reveals three critical gaps: 
\textbf{(1) Virtual-physical ambiguity}. Standard VLMs assume that every object in an image is a physical part of the scene, and are trained on such natural images. However, SAR scenes are  composite of the projected virtual content and physical environment. Without a mechanism to separate the two layers, VLMs suffer from virtual-physical ambiguity, resulting in ``merged" descriptions. For example, a VLM may misidentify the projected ``surfer riding a wave" as a printed picture in the physical scene (see the first example in the blue box of~\cref{fig:teaser}). This confusion prevents the system from making intelligent, context-aware decisions about the SAR scene.
\textbf{(2) Perceptual degradation due to projection}. In SAR, the projected content suffers from various geometric and photometric distortions, such as non-ideal viewpoints, environment light, surface materials and complex shapes. These factors degrade image quality to the point where standard VLM captioning becomes unreliable, and causing hallucinated captions.
\textbf{(3) Lack of SAR semantic benchmark dataset}. The SAR technical community has traditionally focused on low-level tasks, such as tracking and localization~\cite{Kagami2019AnimatedStickies, Nomoto2022dpm, peng2025fp, erel2024casperdpm}, geometric calibration~\cite{huang2021flexible, dong2023Calibration}, shape reconstruction~\cite{huang2021deprocams, Urakawa2025neural} or projector compensation~\cite{grundhofer2015robust, huang2022CompenNeSt++, Kageyama2022deblur, Deng2025GS-ProCams}. There is a lack of large-scale, semantically annotated dataset required to train and evaluate models to interpret the complex SAR scenes. Furthermore, traditional evaluation metrics fail to semantically distinguish between the physical scene and the projected content. While standard semantic evaluation metrics (like BLEU or CIDEr) provide a single score for the whole image, they cannot independently measure if a VLM correctly identified the physical scene but failed to understand the projection, or vice versa.
To address these issues, we propose ProCap. Rather than treating the SAR scene as a single layer, ProCap explicitly models it as a composition of a physical scene and a projection overlay. 
First, to resolve virtual-physical ambiguity, we design an automatic segmentation module to  identify and isolate the projected content from the physical scene using a binary mask. This provides the VLM~\cite{radford2021clip, blip2Qformer} with unambiguous spatial context, avoiding ``merged" descriptions. 
Second, to deal with perceptual degradation, we propose a region-aware retrieval mechanism to extract features from the distorted projection regions, and matches them to a high-fidelity external semantic knowledge base. By retrieving the ``clean" object names for the projected content, ProCap provides the VLM with a strong semantic signal that remains robust against geometric or photometric distortions. 
Finally, to address the lack of SAR semantic benchmark dataset, we introduce the RGBP dataset (RGB + Projections), a large-scale dataset with dual-captioning evaluation protocol specifically created for SAR. Unlike conventional datasets, it provides rich annotations, including precise segmentation masks and separate ground truth (GT) captions for both the physical scene and the projected content. The proposed dual-captioning evaluation protocol particularly assesses the model's ability to generate accurate captions for the physical scene and the projection independently. This prevents high performance in one task and failures in the other. 
Our contributions can be summarized as follows:
\begin{itemize} 
    \item \textbf{ProCap framework.} Our ProCap presents a two-stage pipeline that resolves virtual-physical ambiguity by decoupling projected content from physical scenes through automated segmentation and region-aware semantic retrieval.
    \item \textbf{The RGBP dataset.} We introduce the first large-scale SAR semantic benchmark dataset, featuring 65 diverse physical scenes and 180,000+ projections. It provides decoupled ground truth captions and masks, moving SAR datasets from low-level calibration/correction to high-level semantic understanding.
    \item \textbf{Dual-captioning evaluation protocol.} We establish a novel evaluation protocol using task-specific tokens. This enables independent assessment of scene and projection description, preventing metric bias due to context confusion in complex SAR scenarios.
\end{itemize}

\section{Related Work}\label{sec:related_work}

\subsection{Spatial augmented reality}

Spatial augmented reality (SAR)/Projection mapping (PM)~\cite{kaminokado2019augment, Nomoto2020dpm, Nomoto2020multiproj, Nomoto2022dpm, yasui2024pm, peng2025fp, iwai2025pm} projects virtual contents onto real object surfaces using projector-camera systems, and is widely utilized in creative arts~\cite{erel2023neuralprojectionmapping, erel2024casperdpm, Deng2025LAPIG, Deng2025GS-ProCams}, industrial design~\cite{grundhofer2015robust, huang2021flexible, dong2023Calibration}, and entertainment~\cite{iwai2006limpiddesk, bermano2017makeuplamps, Kagami2019AnimatedStickies}. 
Interactive and dynamic PM extends SAR beyond static visual display toward the manipulation of human perception and real-time interaction. 
Early interactive PM focused on enabling user access to digital information overlaid on physical environments~\cite{iwai2006limpiddesk}.
As interaction increasingly involved moving and deformable targets, subsequent research evolved toward dynamic PM, where the projection target has continuous motion or non-rigid deformation.
Kagami \etal~\cite{Kagami2019AnimatedStickies} achieved robust alignment on moving planar surfaces through a high-speed closed-loop tracking approach.
Further examples include dynamic projection onto human faces~\cite{bermano2017makeuplamps, peng2025fp} and arms~\cite{Peng2020high} at low latency, and highly non-rigid hands~\cite{erel2024casperdpm} combining 3D pose estimation with 2D correction.
In parallel, dynamic PM has advanced in multi-projector configurations, providing larger coverage and adaptive shadow removal in complex dynamic scenes~\cite{Nomoto2020multiproj, Nomoto2022dpm}. To enhance visual realism, high-speed rendering pipelines, such as path tracing~\cite{Nomoto2020dpm}, have been applied to maintain visual coherence at high frame rates.
As diffusion models advance, instructional natural language is been fully utilized to control projected content~\cite{erel2023neuralprojectionmapping, erel2024casperdpm, Deng2025LAPIG}. Deng \etal~\cite{Deng2025LAPIG} proposed language-guided projector image generation method with surface adaptation and stylization.
 
\subsubsection{Virtual-physical ambiguity in SAR}
Virtual-physical ambiguity describes the perceptual blending or confusion of physical scene and projected content in SAR, leading users to perceive it as real~\cite{takeuchi2024projlight}.
To improve it, one solution is projector compensation, which aims to produce the viewer's desired projection effects by modifying the projector input patterns to eliminate the distortions influenced by the environment~\cite{yasui2024pm} and projected surface or object attribute. 
Early methods focused on geometry~\cite{Raskar2003iLamps} and color distortions~\cite{grundhofer2015robust},
yet limited projector's physical brightness may cause over-compensation and artifacts.
To achieve visually satisfactory projections, properties of the human visual system such as chromatic adaptation and perceptual anchoring are leveraged~\cite{huang2017radiometriccompensation}. With the advent of deep learning, photometric~\cite{huang2019compennet} and geometric~\cite{huang2022CompenNeSt++, huang2021deprocams, wang2023CompenHR, wang2024vicomp} compensation are formulated as an end-to-end learning problem.
Leveraging optical illusions~\cite{luo2021staypositive} to produce high-quality non-negative images explores compensation artifacts reduction. 
To handle complex lighting and scene decomposition, ProCams leverages path tracing-based differentiable rendering~\cite{Li2025DPCS} for improved dark texture compensation, while further projection artifacts (\eg, blur and occlusion) are mitigated via learning-based deblurring~\cite{Kageyama2022deblur} and optical system design~\cite{Kusuyama2024deblur}. 
However, a disconnect remains between the virtual-physical ambiguity in SAR and VLM-based reasoning, causing a mismatch with human perceptual expectations. To our best knowledge, no existing method addresses SAR scene understanding via VLMs.

\subsection{Domain-aware vision language model}

Vision language models (VLMs)~\cite{radford2021clip, li2022blip, llama3modelcard, fastvlm2025, zhu2025internvl3exploringadvancedtraining, bai2025qwen25vltechnicalreport, bai2025qwen3vltechnicalreport} aim to connect the two different modalities of vision and language and learn the correspondence between them.
Early VLMs achieve contrastive pre-training on large-scale image-text data~\cite{jia2021scaling, radford2021clip}, with strong cross-domain generalization, laying the foundation for subsequent training paradigms~\cite{alayrac2022flamingo, dai2023instructblip, Wang2023Learning} and generalized visual language tasks (\eg~image retrieval, text generation, \etc)~\cite{fastvlm2025, bai2025qwen25vltechnicalreport, bai2025qwen3vltechnicalreport, zhu2025internvl3exploringadvancedtraining}.
The versatility of VLM has also inspired preliminary explorations into their integration with XR~\cite{xiu2025viddar, duan2025advancing}, motivated by challenges including virtual-physical ambiguity in XR scenes and the difficulty of recognizing seamlessly integrated content. In SAR, several benchmarks were developed for low-level tasks~\cite{huang2019compennet, huang2021deprocams, huang2022CompenNeSt++, erel2023neuralprojectionmapping}. While they have substantially advanced SAR low-level tasks (\eg., compensation), they were limited in rich semantic annotations in SAR, and these challenges are not well addressed by VLMs trained on conventional image-text data. 
Despite the excellent performance of generic models, directly using them for specialized domains suffers from limited effectiveness. To address this issue, solutions have been proposed for adaptation to specialized domains:

\vspace{1mm}\noindent\textbf{Full fine-tuning} is commonly used for VLM domain adaptation. Researchers use small-scale labeled datasets from specific domains to further train the pre-trained general VLMs to adjust model parameters to better fit the tasks and data distribution of specialized domains. However, full fine-tuning of all model parameters is computationally expensive and often leads to overfitting, especially when only limited labeled data is available. To mitigate this, an intermediate step known as domain-specific continued pre-training is often introduced~\cite{gururangan2020don, blip2Qformer}. In this approach, large-scale unlabeled or weakly labeled data from the target domain is used to continuously pre-train the model before downstream task fine-tuning. This allows the model to acquire domain-relevant visual features and linguistic patterns, thereby providing a more suitable initialization for subsequent fine-tuning and improving its effectiveness.

\vspace{1mm}\noindent\textbf{Parameter-Efficient Fine-tuning (PEFT)} methods have been developed to further address the inefficiencies of full fine-tuning. Techniques such as adapter tuning~\cite{houlsby2019parameter} and low-rank adaptation (LoRA)~\cite{hu2022lora} enable efficient adaptation by modifying only a small subset of parameters, significantly reducing computational cost while maintaining competitive performance. These methods add only a small number of trainable parameters or modules to the model while freezing most of the pre-trained model parameters. This not only greatly reduces the computational resources required for training, but also effectively avoids the problem of catastrophic forgetting on general knowledge.

\vspace{1mm}\noindent\textbf{Retrieval Augmentation (RA)} addresses specialized domains with precise knowledge (\eg, legal, medical). Some research efforts have attempted to combine language models with an external knowledge base~\cite{izacard2021leveraging, lewis2020retrieval, Yasunaga2023retrieval, li2024evcap}.
When the model needs to answer a specialized question, it first retrieves relevant documents or cases from the knowledge base, and then uses the retrieved information as a context to generate an answer together with the input visual information.

In summary, although SAR research focuses on projection quality and plausibility illusion, and VLM studies explore domain adaptation and retrieval augmentation, none address this unique challenge of SAR scene understanding. The main obstacle is the lack of benchmarks, as existing datasets (\eg, COCO~\cite{lin2014coco}, nocaps~\cite{Agrawal2019nocaps}, WHOOPS!~\cite{Bitton-Guetta2023WHOOPS}) do not capture projection-related phenomena.

\section{RGBP Dataset}\label{sec:rgbp_dataset}

Existing VLMs are trained on natural images. While datasets like COCO~\cite{lin2014coco} are standard for general recognition, they do not account for the interaction between physical scenes and projected content common in SAR scenes. To address this, we developed the RGBP dataset to train and evaluate VLMs in complex SAR scenes.

\subsection{Data acquisition} 
\begin{figure}[tbp]
  \centering
  \includegraphics[width=0.40\textwidth]{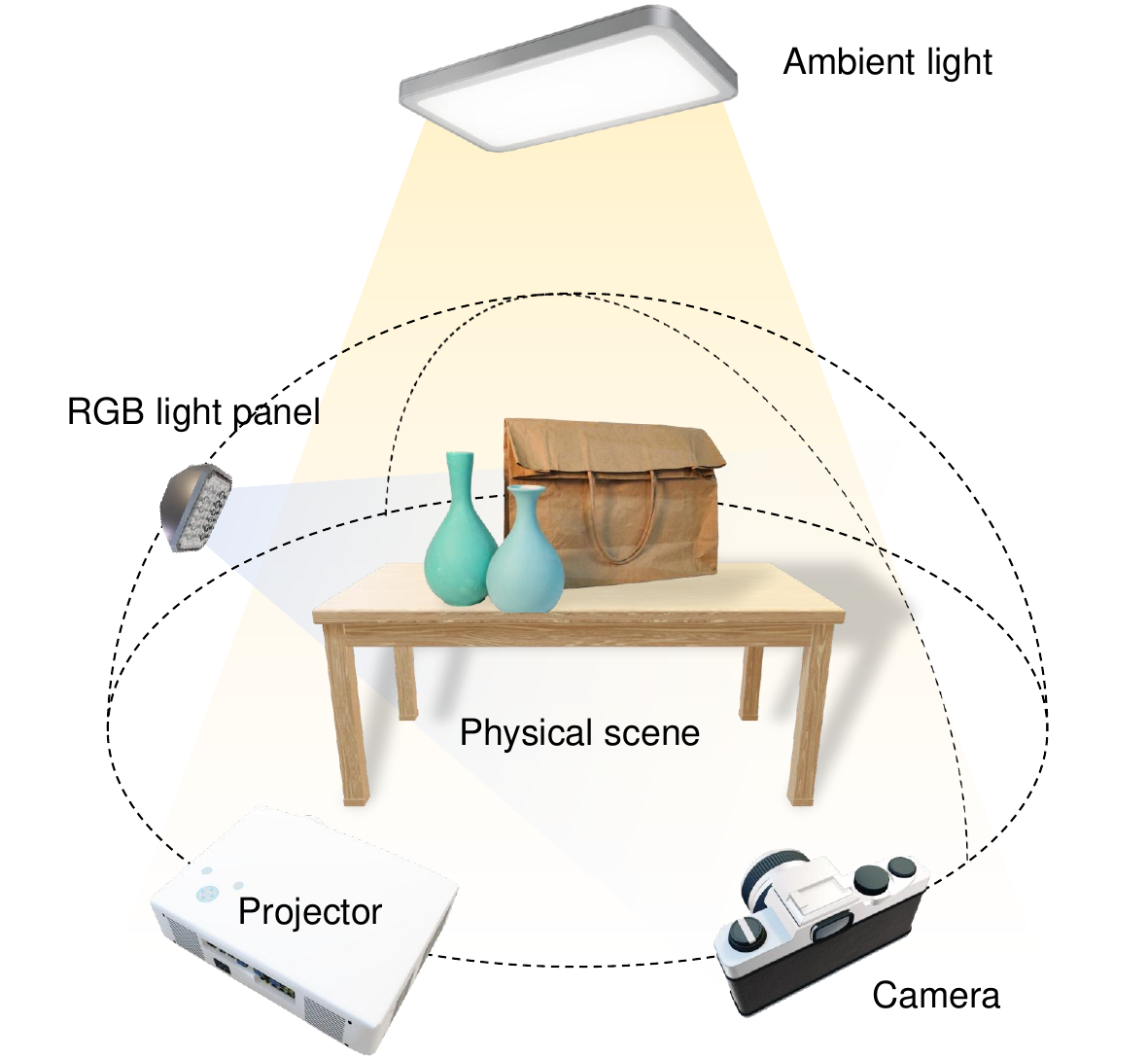}
  \caption{Configuration of the RGBP dataset capture environment.}
  \label{fig:rgbp_capture_environment}
\end{figure}

We used a projector-camera system (ProCams) in a hemispherical setup (\cref{fig:rgbp_capture_environment}). We varied two main factors to ensure the dataset covers a wide range of real-world conditions:

\textbf{Lighting}: We used ambient light and an RGB light panel (YONGNUO YN300Air II) to create scenarios ranging from bright, uniform illumination to dim, high-noise environments. Specifically, the ambient illumination level varied from $\sim$2.9 to 546.0 lux. The RGB light panel was operated in both high and low power modes, producing luminous fluxes of 2,268 lm and 1,235 lm, respectively. 
Projection distortions and occlusions naturally arise from non-planar geometries and surface self-occlusion, particularly in scenes with curved surfaces.

\textbf{Geometry}: We changed the projector's angle and placed physical objects stochastically to create diverse geometric distortions and surface occlusions. The surface shapes in all seen and unseen scenes are explicitly categorized into planar, mildly curved, and highly curved surfaces based on the projection field of view (FOV), as shown in~\cref{tab:rgbp_configuration_surface_curvature_variations}. 

\begin{table}[!tbh]
\centering
\caption{Surface shape variations of the RGBP dataset}
\label{tab:rgbp_configuration_surface_curvature_variations}
\setlength{\tabcolsep}{3pt} 
\begin{tabular}{>{\raggedright\arraybackslash}p{0.12\textwidth} >{\raggedright\arraybackslash}p{0.24\textwidth} >{\raggedright\arraybackslash}p{0.08\textwidth}}
\toprule
\textbf{Surface shape} & \textbf{Seen scenes} & \textbf{Unseen scenes} \\
\midrule
Planar & 01, 02, 03, 04, 05, 06, 07, 08, 09, 21, 25, 28, 29, 37, 45, 47, 60 & 61, 63 \\ \addlinespace
Mildly curved & 10, 11, 12, 13, 15, 17, 18, 19, 20, 23, 24, 32, 33, 34, 35, 39, 46, 49, 51, 52, 53, 55, 56 & 64, 65 \\ \addlinespace
Highly curved & 14, 16, 22, 26, 27, 30, 31, 36, 38, 40, 41, 42, 43, 44, 48, 50, 54, 57, 58, 59 & 62 \\

\bottomrule
\end{tabular}
\end{table}

\subsection{Composition of the training and evaluation dataset}
The dataset composition are shown in~\cref{tab:dataset_composition_detailed}, including hardware and resolution details. The projection image set pulls from COCO 2017~\cite{lin2014coco}, nocaps~\cite{Agrawal2019nocaps}, and WHOOPS!~\cite{Bitton-Guetta2023WHOOPS} for semantic variety.

\begin{figure}[tbp]
  \includegraphics[width=0.48\textwidth]{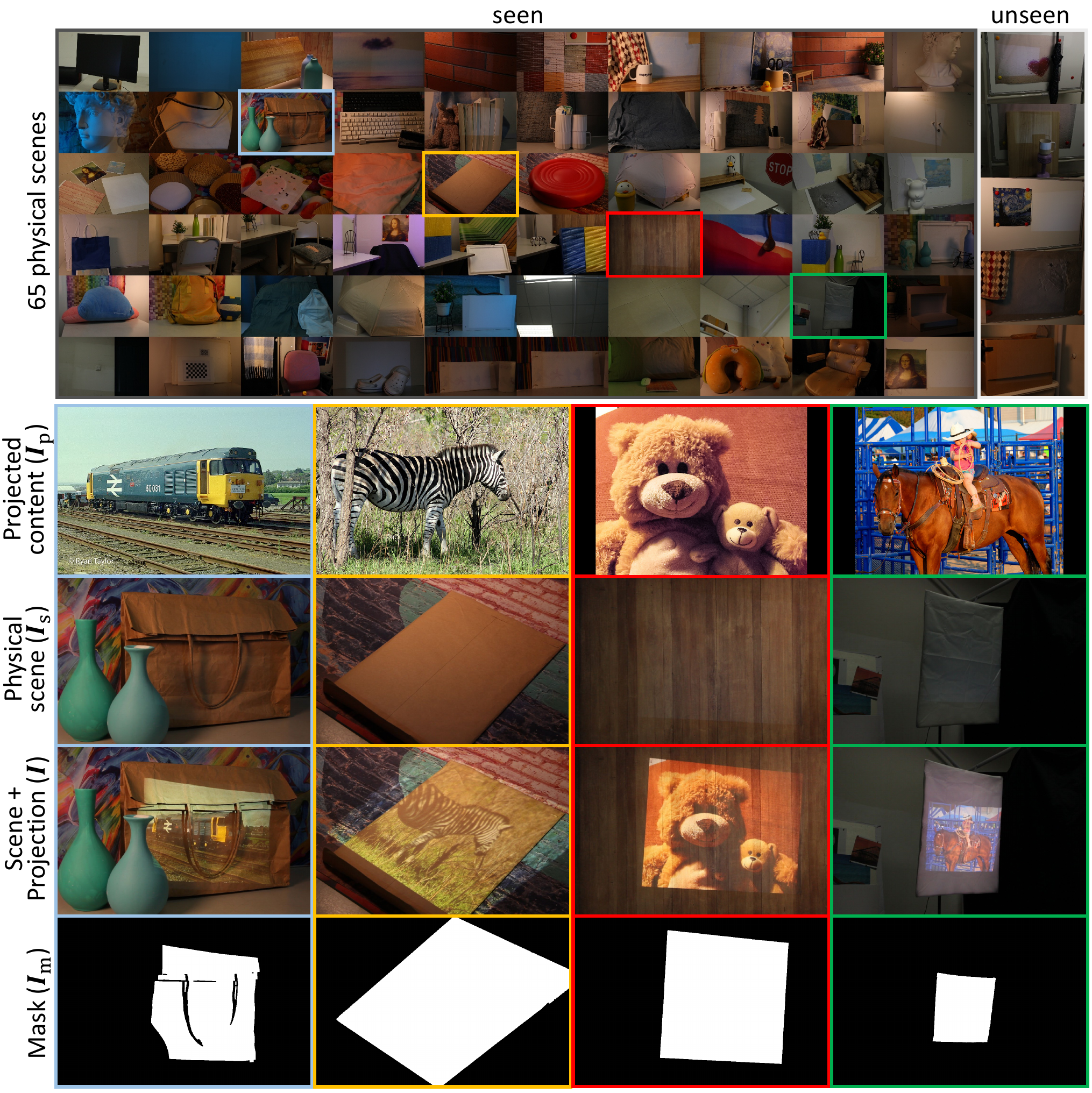}
  \caption{RGBP scenes used to train and evaluate models. We show four representative scenes with projection, highlighted by the blue, yellow, red, and green boxes, respectively. Note that the projection masks are coarse and they do not full match the real projection regions (see \cref{sec:projection_segmentation} for details).}
  \label{fig:rgbp_trained_scene}
\end{figure}

\textbf{Training set}. We used 118,287 images from the COCO train split, distributed across 60 physical scenes. Each scene includes 2,000 \textbf{unique} projections (287 for the final scene, since scene60 only has 287 images). To help the model separate the physical scene from the projection, we also captured baseline frames for each scene using pure black, white, and gray projections. Then, we leveraged language models to assist in generating 50 high quality captions per scene. Five of these captions were randomly matched to each individual image. This strategy enhances both the diversity and accuracy of physical scene captions. Some training samples are shown in~\cref{fig:rgbp_trained_scene}.

\textbf{Evaluation set}. The evaluation set includes 62,400 image pairs across 60 seen and 5 unseen scenes, using 960 test images from COCO val, nocaps val, and WHOOPS! used as unseen projected content. We also captured five additional scenes using novel ProCams in supplementary material.

\begin{table}[t]
\centering
\caption{Composition of the proposed RGBP dataset. Training data are collected from 60 physical scenes using COCO train images as projected content; $^*$one COCO scene contains only 287 images. The evaluation set comprises 65 physical scenes, including 60 seen and 5 unseen, using COCO validation, nocaps, and WHOOPS! images as projected content. Total pairs = physical scenes $\times$ projected images per scene. }
\label{tab:dataset_composition_detailed}
\resizebox{\columnwidth}{!}{
\begin{tabular}{@{}lcrr@{}}
\toprule
\textbf{Split} & \textbf{Physical scenes} & \textbf{Projected images per scene} & \textbf{Total pairs} \\ \midrule
Train & 60 & COCO train (2,000$^*$) & \textbf{118,287} \\ \midrule
 &  & COCO val (160) &  \\
Eval. & 60 (seen) + 5 (unseen) & nocaps val (300) & \textbf{62,400} \\
 &  & WHOOPS! (500) & \\ \midrule
\textbf{Total} & \textbf{65} &  & \textbf{180,678}\\ \bottomrule
\end{tabular}
}
\end{table}

\subsection{Dual-captioning evaluation protocol}
Instead of a single caption, RGBP uses a dual-captioning annotation scheme. For every image, we provide: (1) A coarse binary segmentation mask of the projected region; (2) Two separate ground truth (GT) captions, \ie, a Scene GT (describing the room/objects) and a Projection GT (describing the projected content).

A key innovation of the RGBP dataset is its specialized annotation scheme, which enables a granular, decoupled evaluation. For each generated image pair, we provide not only a precise binary segmentation mask but also two distinct GT captions, which form the basis of our dual-captioning evaluation protocol.

During evaluation, the model is prompted to generate two distinct captions. This allows us to measure how well the model understands the physical scenes versus the projected content independently, providing a clearer picture of its performance than a single holistic score in SAR applications.

\section{Projection-Aware Captioning (ProCap)}\label{sec:method}

\begin{figure*}[t]
    \centering
    \includegraphics[width=\textwidth]{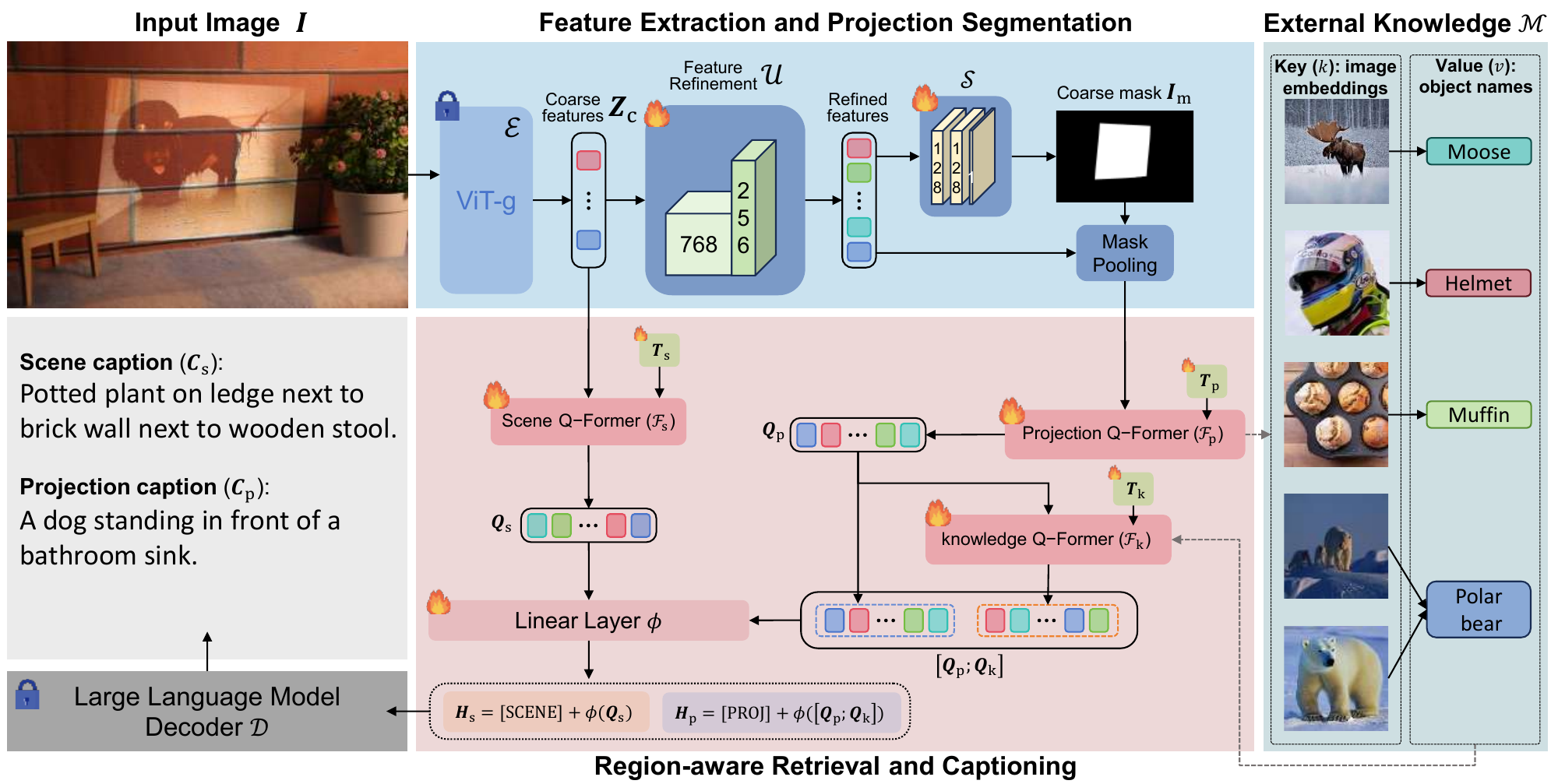}
    \caption{Overview of the proposed \textbf{ProCap} architecture. Given an observed image $\I$ containing both physical scene and projected content, a frozen vision transformer (ViT-g) backbone first extracts coarse features $\boldsymbol{Z}_\text{c}$, which are refined into $\mathcal{U}(\boldsymbol{Z}_\text{c})$ by a feature refinement module $\mathcal{U(\cdot)}$. A projection segmentation module $\mathcal{S}$ is employed to estimate a coarse projection mask $\Imask$, enabling mask pooling to retain projection features as $\boldsymbol{Z}_\text{p}$. The scene and projection features $\boldsymbol{Z}_\text{c}$ and $\boldsymbol{Z}_\text{p}$ are then processed by two specialized Q-Formers: a \textbf{scene Q-Former} and a \textbf{projection Q-Former} to obtain scene embeddings $\boldsymbol{Q}_\text{s}$ and projection embeddings $\boldsymbol{Q}_\text{p}$, respectively. $\boldsymbol{Q}_\text{p}$ is further used to retrieve similar object names (semantic context) $\boldsymbol{N}$ from external semantic knowledge base $\mathcal{M}$. The retrieved semantic context and $\boldsymbol{Q}_\text{p}$ together are then encoded by a \textbf{knowledge Q-Former} as $\boldsymbol{Q}_\text{k}$. Finally, the scene embeddings $\boldsymbol{Q}_\text{s}$, projection embeddings $\boldsymbol{Q}_\text{p}$, and semantic context embeddings $\boldsymbol{Q}_\text{k}$ are projected into the embedding space of a frozen LLM decoder via a linear layer $\phi$, conditioned to generate separate captions for both the physical scene and the projected content. Lock and fire symbols stand for frozen and trainable parameters, respectively.
}
\label{fig:procap_architecture}
\end{figure*}

\subsection{Problem formulation}
In an SAR environment, the observed image $\I$ (the fourth row of \cref{fig:rgbp_trained_scene}) is a composite of the physical scene (the second row of \cref{fig:rgbp_trained_scene}) and the projected content (the third row of \cref{fig:rgbp_trained_scene}). We model this image formation as:
\begin{equation}
\I = \I_{\text{s}} \oplus \I_{\text{p}},
\end{equation}
where $\I_{\text{s}}$ represents the physical scene, $\I_{\text{p}}$ denotes the projected content, and $\oplus$ represents the complex photometric and geometric blending between them. This blending is influenced by surface shape and material, ambient lighting, and projector-camera geometry, making $\I_{\text{p}}$ inherently distorted compared to the original digital image.

The objective of ProCap is to resolve the virtual-physical ambiguity by recovering independent semantic descriptions for both layers. Inspired by EVCap~\cite{li2024evcap}, we formulate this as a dual-captioning task:
\begin{equation}
(\boldsymbol{C}_{\text{s}}, \boldsymbol{C}_{\text{p}}) = \mathcal{F}(\I),
\end{equation}
where $\boldsymbol{C}_{\text{s}}$ and $\boldsymbol{C}_{\text{p}}$ are natural language descriptions of the physical scene and the projected content, respectively.

Unlike standard VLMs that treat $\I$ as a single semantic layer, which may lead to ``merged" descriptions. 
ProCap decomposes the task into two stages: We first segment the composite image $\I$ to isolate the projection regions dominated by $\I_{\text{p}}$ from those representing $\I_{\text{s}}$ (\cref{{sec:projection_segmentation}}). Then, we extract features from the distorted projection regions and matching them against a high-fidelity knowledge base to avoid ambiguous semantic context of $\boldsymbol{C}_{\text{p}}$ due to geometric and photometric distortion (\cref{{sec:region-aware_retrieval_and_captioning}}). 

\subsection{Feature extraction and projection segmentation}\label{sec:projection_segmentation}
The first stage of ProCap isolates the projected content from the physical scene.
We employ a frozen CLIP ViT-g \cite{sun2023EVA-CLIP} as our vision encoder $\mathcal{E}$ to map the input image $\I$ into a coarse feature map $\boldsymbol{Z}_\text{c}$. While $\boldsymbol{Z}_\text{c}$ captures global context, its resolution is often insufficient for small or distorted projected regions. To resolve this, we apply two deconvolutional layers $\mathcal{U}$ to upsample the features. Then, we apply a two-layer convolutional network-based segmentation module $\mathcal{S}$ for this purpose. Given an input image $\I$, the module generates a coarse binary mask $\Imask$: 
\begin{equation}
\Imask = \mathcal{S}(\mathcal{U}(\boldsymbol{Z}_\text{c})),
\end{equation}
where $\Imask(x,y) = 1$ denotes the presence of projected light, and $\Imask(x,y) = 0$ represents the unaugmented physical scene. To ensure computational robustness in varied lighting conditions, we employ a coarse masking strategy rather than attempting to recover high-frequency instance boundaries. We derive the target mask from a reference white-light projection for each scene. This approach serves two purposes: First, it defines the addressable area of the projector, effectively acting as a spatial regularizer that ignores peripheral boundary noise. Second, it prevents the model from overfitting to the specific textures or colors of the projected content. By focusing on the central region of the projection, we provide a stable spatial prior that facilitates unambiguous feature extraction in the subsequent retrieval stage.

\subsection{Region-aware retrieval and captioning}\label{sec:region-aware_retrieval_and_captioning}
With the coarse projection mask $\Imask$ obtained, ProCap extracts and enhances visual features to generate dual descriptions. This process is designed to overcome the perceptual degradation due to projection by anchoring distorted visual signals to a clean semantic knowledge base.

\subsubsection{Feature decoupling and mask pooling}
We apply a Mask Pooling operation to combine the upsampled features $\mathcal{U}(\boldsymbol{Z}_\text{c})$ with the mask $\Imask$, producing the projection-specific features $\boldsymbol{Z}_\text{p}$:
\begin{equation}
\boldsymbol{Z}_\text{p} = \text{MaskPool}\left(\mathcal{U}(\boldsymbol{Z}_\text{c}), \Imask\right).
\end{equation}
We then use two sets of learnable query tokens, $\boldsymbol{T}_\text{s}$ and $\boldsymbol{T}_\text{p}$, within a Q-Former~\cite{blip2Qformer} architecture $\mathcal{F}$ to produce fixed-length embeddings for the physical scene ($\boldsymbol{Q}_\text{s}$) and the projected content ($\boldsymbol{Q}_\text{p}$):
\begin{equation}
\boldsymbol{Q}_\text{s} = \mathcal{F}_\text{s}(\boldsymbol{Z}_\text{c}, \boldsymbol{T}_\text{s}), \quad
\boldsymbol{Q}_\text{p} = \mathcal{F}_\text{p}(\boldsymbol{Z}_\text{p}, \boldsymbol{T}_\text{p}).
\end{equation}

\subsubsection{Semantic context retrieval via external knowledge}
To deal with the impact of photometric and geometric distortions, we augment the projection features with an external knowledge base (or visual-name memory~\cite{li2024evcap}) $\mathcal{M}$. This knowledge base is a key-value store $\mathcal{M} = \{(k_i, v_i)\}_{i=1}^n$ derived from the LVIS dataset~\cite{gupta2019lvis}, where $k_i$ is a visual embedding and $v_i$ is its textual label. We retrieve the top-$K$ ($K=9$ in our implementation) most similar object names $\boldsymbol{N}$ by computing the cosine similarity between $\boldsymbol{Q}_\text{p}$ and the keys in $\mathcal{M}$. To filter noise and redundant information from $\boldsymbol{N}$, a knowledge Q-Former distills these names into a compact semantic context embeddings  $\boldsymbol{Q}_\text{k}$: 
\begin{equation}
\boldsymbol{Q}_\text{k} = \mathcal{F}_\text{k}(\boldsymbol{N}, \boldsymbol{Q}_\text{p}, \boldsymbol{T}_\text{k}), \label{eq:qn}
\end{equation}
where $\boldsymbol{T}_\text{k}$ is a learnable semantic query token. This retrieval step effectively replaces noisy projected pixels with clean semantic priors, significantly improving captioning accuracy.

\subsubsection{Multi-source fusion and dual-captioning}
To prepare the features for the dual-captioning LLM, we project the embeddings (\cref{{eq:qn}}) into the LLM's input space using a linear layer $\phi$. The final prompt $\boldsymbol{H}$ is constructed by:
\begin{equation}
\boldsymbol{H}_\text{s} = \text{[SCENE]} + \phi(\boldsymbol{Q}_\text{s}), \quad \boldsymbol{H}_\text{p} = \text{[PROJ]} + \phi([\boldsymbol{Q}_\text{p}; \boldsymbol{Q}_\text{k}]).
\end{equation}
To ensure the model distinguishes between the two tasks, we prepend task-specific tokens (\eg, [SCENE] and [PROJ]) to the generation sequence. Operator $[\cdot\ ;\ \cdot ]$ stands for concatenation. Finally, a frozen LLM decoder $\mathcal{D}$ generates the captions $\boldsymbol{C}_{\text{s}} = \mathcal{D}(\boldsymbol{H}_\text{s})$ and $\boldsymbol{C}_{\text{p}}=\mathcal{D}(\boldsymbol{H}_\text{p})$ autoregressively. 

\subsection{Training losses}
We train the trainable parameters $\theta$ (the refinement layers $\mathcal{U}$, segmentation module $\mathcal{S}$, query tokens $\{\boldsymbol{T}_\text{s}, \boldsymbol{T}_\text{p}, \boldsymbol{T}_\text{k}\}$, linear projection $\phi$, and the cross-attention layers within the Q-Formers) via an end-to-end multi-task objective that jointly minimizes projection segmentation and captioning losses. The total loss function $\mathcal{L}$ is defined as a weighted sum of the scene captioning loss $\mathcal{L}_{\text{s}}$, the projection captioning loss $\mathcal{L}_{\text{p}}$, and the projection segmentation loss $\mathcal{L}_{\text{seg}}$:
\begin{equation}
\mathcal{L}(\theta) = \alpha \mathcal{L}_\text{s} + \beta \mathcal{L}_\text{p} + \gamma,  \mathcal{L}_\text{seg}
\end{equation}
where hyperparameters $\alpha = 0.5$, $\beta = 0.5$, and $\gamma = 1.0$.

For the dual-captioning tasks, the objective is to minimize the negative log-likelihood of the ground truth caption sequence $(c_1, ..., c_L)$ for both the physical scene ($\boldsymbol{C}_{\text{s}}$) and projected content ($\boldsymbol{C}_{\text{p}}$):
\begin{align}
\mathcal{L}_\text{s} = -\sum_{t=1}^{L} \log P(c_t \mid \boldsymbol{H}_\text{s}, c_{1:t-1}; \theta) \\
\mathcal{L}_\text{p} = -\sum_{t=1}^{L} \log P(c_t \mid \boldsymbol{H}_\text{p}, c_{1:t-1}; \theta)
\end{align}
We also apply binary cross-entropy (BCE) loss between the predicted mask $\hat{\boldsymbol{I}}_\text{m}$ and the ground truth mask $\Imask$ to supervise accurate projection mask segmentation:
\begin{equation}
\mathcal{L}_\text{seg} = \text{BCE}(\hat{\boldsymbol{I}}_\text{m}, \Imask).
\end{equation}

\begin{figure*}[!tb]
    \centering
    \includegraphics[width=\textwidth]{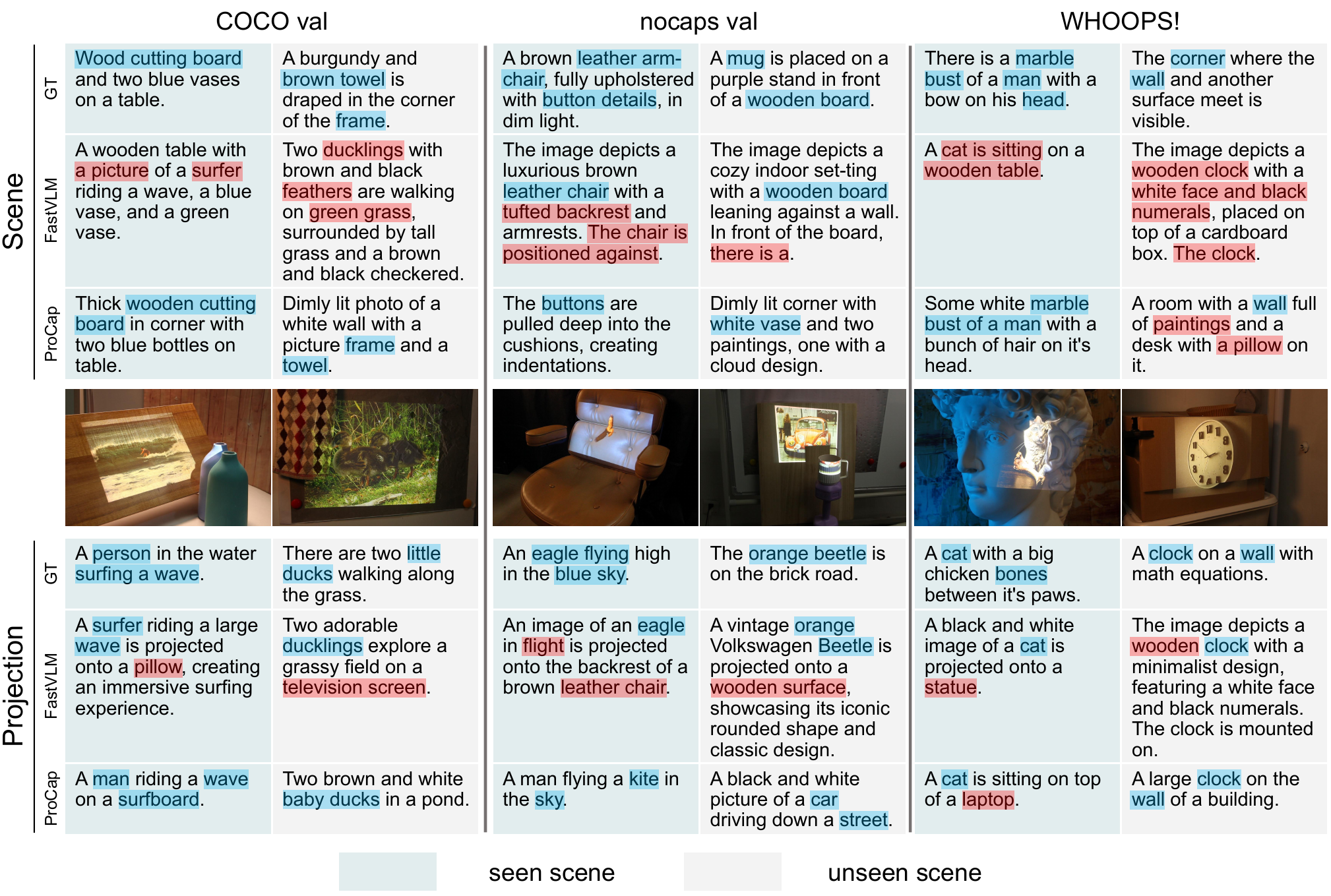}
    \caption{Qualitative comparison of descriptive ability in complex SAR scenes. We evaluate ProCap~$_{\text{OpenLLaMA-3B}}$ against FastVLM-7B~\cite{fastvlm2025} across three seen and three unseen scenes. The top panels compare descriptions of physical scenes, while the bottom panels focus on projected content. Incorrect descriptions are highlighted in \false{red} and correct ones in \true{blue}. Note that the model outputs are truncated by the max tokens limit.}
    \label{fig:rgbp_qualitative_results}
\end{figure*}

\section{Experiments}\label{sec:experiments}

\subsection{System configuration}
We evaluated ProCap using several open-source LLMs as decoders ($\mathcal{D}$) to demonstrate architectural flexibility: TinyLlama-1.1B~\cite{zhang2024tinyllama}, OPT-2.7B~\cite{zhang2022optopenpretrainedtransformer}, OpenLLaMA-3B~\cite{touvron2023llama}, Vicuna-1.5-7B~\cite{zheng2023judgingllmasajudgemtbenchchatbot} for diversity. In addition, we fine-tuned Qwen3-VL-8B-Instruct~\cite{bai2025qwen3vltechnicalreport} on our RGBP dataset using 
Supervised Fine-Tuning (SFT) and LoRA, following the procedure
in~\cite{Qwen2-VL-Finetuning}. Key training parameters were kept consistent across experiments: Training and inference were conducted on one NVIDIA A100, one NVIDIA RTX PRO 6000 Blackwell, and three NVIDIA RTX 4060Ti GPUs. More details are in the supplementary.

\vspace{0.5em}\noindent\textbf{Baselines and metrics} We compared ProCap against state-of-the-art efficient VLMs: FastVLM (0.5B, 1.5B, and 7B)~\cite{fastvlm2025} and Qwen3-VL-Instruct (2B, 4B, and 8B)~\cite{bai2025qwen3vltechnicalreport}. For a fair comparison, baselines were prompted with detailed instructions to perform the same dual-captioning description:

\noindent
\fcolorbox{black}{gray!10}{
\begin{minipage}{0.98\dimexpr\linewidth-2\fboxsep-2\fboxrule\relax}
\small\ttfamily
\#\# \textsl{For scene} \\
Describe the scene in detail in the image, excluding any projected content, as a short image caption. \\
\#\# \textsl{For projection} \\
Describe any projected content in detail in the image, excluding the surrounding scene, as a short image caption.
\end{minipage}
}

Performance was measured using BLEU@4~\cite{Papineni2002bleu}, METEOR~\cite{banerjee2005meteor}, CIDEr (C)~\cite{Vedantam2015CIDEr}, and SPICE (S)~\cite{spice2016}. 
BLEU and METEOR focus on surface-level n-gram overlap and linguistic variations, but limited in complex semantic nuances. SPICE addresses this by leveraging scene graphs to evaluate propositional content. CIDEr utilizes term frequency-inverse document frequency (TF-IDF) to weight informative terms, further aligns more closely with human consensus. We prioritize CIDEr and SPICE in SAR to ensure that the model's understanding remains consistent with human perceptual standards.

To independently evaluate scene and projection understanding, we introduce a dual-captioning evaluation protocol, which compares the model's generated descriptions for the physical scene and the projected content against their respective GT captions.

\begin{table}[!tb]
    \centering
    \renewcommand{\arraystretch}{1.2} 
    \caption{Comparison on \textbf{the 60 seen scenes with unseen projected content}. CIDEr (C), SPICE (S) are reported. Results are averaged on all scenes. The best results are in \textbf{bold}, and the second best results are \underline{underlined}. See the supplementary for the full table\protect\footnotemark[1].}
    \label{tab:results_60_seen_scenes}

\resizebox{\linewidth}{!}{
\begin{tabular}{@{}cl rr rr rr}
\toprule[1pt]
\multirow{3}{*}{\textbf{Task}} & \multirow{3}{*}{\textbf{Method}} 
& \multicolumn{2}{c}{\textbf{COCO}} & \multicolumn{2}{c}{\textbf{nocaps val}} & \multicolumn{2}{c}{\textbf{WHOOPS!}} \\
& & \multicolumn{2}{c}{Test} & \multicolumn{2}{c}{Overall} & \multicolumn{2}{c}{Test} \\
\cmidrule(lr){3-4} \cmidrule(lr){5-6} \cmidrule(lr){7-8}
& & \multicolumn{1}{c}{C $\uparrow$} & \multicolumn{1}{c}{S $\uparrow$} & \multicolumn{1}{c}{C $\uparrow$} & \multicolumn{1}{c}{S $\uparrow$} & \multicolumn{1}{c}{C $\uparrow$} & \multicolumn{1}{c}{S $\uparrow$} \\
\midrule
\multirow{5}{*}[-3.8em]{%
  \centering
  \rotatebox[origin=c]{90}{%
  \shortstack[c]{Scene captioning}
  }%
}
& FastVLM-0.5B~\cite{fastvlm2025} & 1.72 & 9.20 & 1.60 & 9.47 & 1.59 & 10.17 \\
& FastVLM-1.5B~\cite{fastvlm2025}  & 1.92 & 8.77 & 1.73 & 9.16 & 1.74 & 9.92 \\
& FastVLM-7B~\cite{fastvlm2025}  & 2.31 & 10.31 & 2.21 & 10.79 & 2.21 & 11.98 \\
& Qwen3-VL-2B-Instruct~\cite{bai2025qwen3vltechnicalreport} & 2.34 & 12.17 & 2.12 & 12.17 & 2.01 & 12.83 \\
& Qwen3-VL-4B-Instruct~\cite{bai2025qwen3vltechnicalreport} & 2.38 & 11.68 & 2.19 & 11.94 & 2.02 & 12.15 \\
& Qwen3-VL-8B-Instruct~\cite{bai2025qwen3vltechnicalreport} & 2.38 & 13.03 & 2.09 & 13.15 & 1.97 & 13.07 \\
& ProCap~{$_{\text{TinyLlama-1.1B}}$ (ours)} & \textbf{70.27} & 21.92 & 35.21 & \underline{24.21} & \textbf{69.95} & 21.87 \\
& ProCap~{$_{\text{OPT-2.7B}}$ (ours)} & 28.01 & \textbf{24.34} & 27.63 & \textbf{24.45} & 28.19 & \textbf{24.11} \\
& ProCap~{$_{\text{OpenLLaMA-3B}}$ (ours)} & \underline{69.43} & 22.01 & \underline{37.14} & 22.75 & \underline{69.46} & 22.08 \\
& ProCap~{$_{\text{Vicuna-1.5-7B}}$ (ours)} & 36.17 & 22.75 & 36.39 & 22.88 & 36.53 & 22.70 \\
& Qwen3-VL-8B-Instruct~{$_{\text{RGBP}}$~\cite{bai2025qwen3vltechnicalreport}} & 37.81 & \underline{23.22} & \textbf{37.48} & 23.49 & 37.70 & \underline{23.59} \\
\midrule
\multirow{5}{*}[-3.6em]{%
  \centering
  \rotatebox[origin=c]{90}{%
  \shortstack[c]{Projection captioning}
  }%
}
& FastVLM-0.5B~\cite{fastvlm2025} & 7.17 & 7.43 & 6.16 & 5.68 & 5.54 & 5.85 \\
& FastVLM-1.5B~\cite{fastvlm2025} & 7.21 & 7.31 & 5.72 & 5.88 & 4.14 & 6.09 \\
& FastVLM-7B~\cite{fastvlm2025} & 7.65 & 7.33 & 7.01 & 6.12 & 6.25 & 6.77 \\
& Qwen3-VL-2B-Instruct~\cite{bai2025qwen3vltechnicalreport} & 10.73 & 12.47 & 14.85 & 10.38 & 9.82 & \underline{11.05} \\
& Qwen3-VL-4B-Instruct~\cite{bai2025qwen3vltechnicalreport} & 10.59 & 11.72 & 14.93 & \underline{10.47} & 10.35 & 10.91 \\
& Qwen3-VL-8B-Instruct~\cite{bai2025qwen3vltechnicalreport} & 11.56 & 11.63 & 15.57 & 10.01 & 11.19 & 10.97 \\
& ProCap~{$_{\text{TinyLlama-1.1B}}$ (ours)} & 54.37 & 9.75 & 30.45 & 4.80 & 11.88 & 3.62 \\
& ProCap~{$_{\text{OPT-2.7B}}$ (ours)} & 57.72 & 10.21 & 30.58 & 5.17 & 15.04 & 4.40 \\
& ProCap~{$_{\text{OpenLLaMA-3B}}$ (ours)} & 76.98 & 13.58 & \underline{42.06} & 6.66 & 19.37 & 5.43 \\
& ProCap~{$_{\text{Vicuna-1.5-7B}}$ (ours)} & \underline{78.99} & \underline{13.83} & 39.93 & 6.59 & \underline{22.33} & 6.02 \\
& Qwen3-VL-8B-Instruct~{$_{\text{RGBP}}$~\cite{bai2025qwen3vltechnicalreport}} & \textbf{127.58} & \textbf{21.14} & \textbf{102.67} & \textbf{13.19} & \textbf{80.46} & \textbf{16.07} \\
\bottomrule[1pt]
\end{tabular}
}
\end{table}

\footnotetext[1]{Qwen3 (-VL) was released after the submission deadline of this work. Following reviewers' suggestions, we additionally report its evaluation and fine-tuning results. However, it was not adopted as the backbone of our framework due to time constraints and will be explored in future work.}

\begin{table*}[t]
    \centering
    \renewcommand{\arraystretch}{1.2} 
    \caption{Comparison on the \textbf{5 unseen scenes with unseen projected content}. BLEU@4 (B@4), METEOR (M), CIDEr (C), SPICE (S) are reported. The best and the second best results are in \textbf{bold} and \underline{underlined}, respectively. See the supplementary for the full table\protect\footnotemark[1].}
    \label{tab:results_5_unseen_scenes}
\resizebox{\linewidth}{!}
{
\begin{tabular}{cl rrrr rrrrrrrr rr}
\toprule[1pt]
\multirow{3}{*}{\textbf{Task}} & \multirow{3}{*}{\textbf{Method}} & \multicolumn{4}{c}{\textbf{COCO}} & \multicolumn{8}{c}{\textbf{nocaps val}} & \multicolumn{2}{c}{\textbf{WHOOPS!}} \\
& & \multicolumn{4}{c}{Test} & \multicolumn{2}{c}{In-domain} & \multicolumn{2}{c}{Near-domain} & \multicolumn{2}{c}{Out-domain} & \multicolumn{2}{c}{Overall} & \multicolumn{2}{c}{Test} \\
\cmidrule(lr){3-6} \cmidrule(lr){7-14} \cmidrule(lr){15-16}
& & \multicolumn{1}{c}{B@4 $\uparrow$} & \multicolumn{1}{c}{M $\uparrow$} & \multicolumn{1}{c}{C $\uparrow$} & \multicolumn{1}{c}{S $\uparrow$} & \multicolumn{1}{c}{C $\uparrow$} & \multicolumn{1}{c}{S $\uparrow$} & \multicolumn{1}{c}{C $\uparrow$} & \multicolumn{1}{c}{S $\uparrow$} & \multicolumn{1}{c}{C $\uparrow$} & \multicolumn{1}{c}{S $\uparrow$} & \multicolumn{1}{c}{C $\uparrow$} & \multicolumn{1}{c}{S $\uparrow$} & \multicolumn{1}{c}{C $\uparrow$} & \multicolumn{1}{c}{S $\uparrow$} \\
\midrule
\multirow{5}{*}[-4em]{%
  \centering
  \rotatebox[origin=c]{90}{%
  \shortstack[c]{Scene captioning}
  }%
}
& FastVLM-0.5B~\cite{fastvlm2025} & 2.16 & 10.98 & 1.44 & 5.42 & 1.42 & 5.20 & 1.50 & 4.96 & 1.90 & 6.62 & 1.32 & 5.58 & 1.02 & 6.78 \\
& FastVLM-1.5B~\cite{fastvlm2025} & 1.88 & 10.90 & 0.98 & 5.36 & 1.14 & 5.14 & 1.14 & 5.98 & 1.48 & 8.16 & 1.04 & 6.44 & 0.84 & 8.50 \\
& FastVLM-7B~\cite{fastvlm2025} & 2.10 & 11.64 & 1.24 & 6.00 & 1.60 & 6.30 & 1.48 & 6.02 & 1.88 & 7.80 & 1.38 & 6.72 & 1.34 & 8.66 \\
& Qwen3-VL-2B-Instruct~\cite{bai2025qwen3vltechnicalreport} & 4.48 & 15.42 & 2.36 & 9.32 & 2.56 & 8.66 & 2.66 & 8.64 & 2.66 & 9.24 & 2.18 & 8.86 & 1.86 & 8.94 \\
& Qwen3-VL-4B-Instruct~\cite{bai2025qwen3vltechnicalreport} & 5.06 & \underline{16.60} & 3.44 & 9.08 & 4.02 & 8.64 & 3.98 & 8.72 & 4.12 & \underline{9.26} & 3.42 & 8.88 & 2.98 & 9.04 \\
& Qwen3-VL-8B-Instruct~\cite{bai2025qwen3vltechnicalreport} & \underline{5.32} & \textbf{17.32} & 3.68 & \textbf{12.26} & 4.12 & \textbf{11.88} & 3.96 & \textbf{11.86} & 4.16 & \textbf{12.32} & 3.52 & \textbf{12.00} & 3.08 & \textbf{12.46} \\
& ProCap~{$_{\text{TinyLlama-1.1B}}$ (ours)} & 2.48 & 11.38 & \underline{5.02} & 5.53 & 4.94 & 8.34 & 4.74 & 8.34 & \textbf{6.08} & 8.50 & 4.06 & 8.34 & \textbf{7.48} & 5.15 \\
& ProCap~{$_{\text{OPT-2.7B}}$ (ours)} & 4.10 & 14.46 & \textbf{5.80} & \underline{9.44} & \textbf{5.78} & \underline{9.06} & \underline{5.62} & \underline{9.24} & 5.50 & 9.06 & \underline{4.82} & \underline{9.12} & 4.68 & 8.90 \\
& ProCap~{$_{\text{OpenLLaMA-3B}}$ (ours)} & 1.50 & 11.73 & 3.78 & 7.77 & 4.88 & 7.16 & 4.96 & 7.28 & 5.56 & 7.14 & 4.02 & 7.10 & \underline{5.73} & 6.60 \\
& ProCap~{$_{\text{Vicuna-1.5-7B}}$ (ours)} & 1.90 & 11.66 & 2.94 & 7.50 & 2.80 & 7.46 & 2.76 & 7.18 & 3.04 & 7.26 & 2.34 & 7.30 & 2.22 & 7.12 \\
& Qwen3-VL-8B-Instruct~{$_{\text{RGBP}}$~\cite{bai2025qwen3vltechnicalreport}} & \textbf{5.64} & 13.64 & 4.82 & 9.40 & \underline{5.76} & 8.84 & \textbf{6.04} & 8.96 & \underline{5.92} & 8.80 & \textbf{4.96} & 8.86 & 4.00 & \underline{9.42} \\
\midrule
\multirow{5}{*}[-3em]{%
  \centering
  \rotatebox[origin=c]{90}{%
  \shortstack[c]{Projection captioning}
  }%
}
& FastVLM-0.5B~\cite{fastvlm2025} & 6.28 & 16.28 & 11.54 & 9.38 & 8.40 & 7.60 & 10.48 & 7.92 & 6.42 & 5.66 & 8.94 & 7.02 & 8.72 & 8.06 \\
& FastVLM-1.5B~\cite{fastvlm2025} & 6.20 & 15.38 & 10.00 & 8.92 & 7.58 & 7.84 & 7.48 & 7.86 & 4.60 & 6.02 & 6.84 & 7.26 & 4.88 & 7.70 \\
& FastVLM-7B~\cite{fastvlm2025} & 6.44 & 16.36 & 9.70 & 9.38 & 8.62 & 8.50 & 9.70 & 9.20 & 8.80 & 6.62 & 9.44 & 8.14 & 7.84 & 9.08 \\
& Qwen3-VL-2B-Instruct~\cite{bai2025qwen3vltechnicalreport} & 10.40 & \underline{22.18} & 14.10 & \underline{16.14} & 17.96 & \underline{13.34} & 21.50 & \underline{13.24} & 19.16 & \underline{12.04} & 19.60 & \underline{12.84} & 13.04 & \underline{13.78} \\
& Qwen3-VL-4B-Instruct~\cite{bai2025qwen3vltechnicalreport} & 6.82 & 20.44 & 12.86 & 14.32 & 17.46 & 12.90 & 20.34 & 13.16 & 17.74 & 11.72 & 18.46 & 12.58 & 12.72 & 12.88 \\
& Qwen3-VL-8B-Instruct~\cite{bai2025qwen3vltechnicalreport} & 7.52 & 20.58 & 14.34 & 14.68 & 18.42 & 12.30 & 21.46 & 13.04 & 18.06 & 11.64 & 19.20 & 12.34 & 13.90 & 13.10 \\
& ProCap~{$_{\text{TinyLlama-1.1B}}$ (ours)} & 13.13 & 17.92 & 46.90 & 12.60 & 42.32 & 6.10 & 33.64 & 5.26 & 19.70 & 3.96 & 32.36 & 5.10 & 15.23 & 4.57 \\
& ProCap~{$_{\text{OPT-2.7B}}$ (ours)} & 19.70 & 17.78 & 63.40 & 10.96 & 51.70 & 7.48 & 32.22 & 5.78 & 16.50 & 3.82 & 34.18 & 5.68 & 16.72 & 4.82 \\
& ProCap~{$_{\text{OpenLLaMA-3B}}$ (ours)} & 19.90 & 21.22 & 66.03 & 14.90 & 58.22 & 8.66 & \underline{48.00} & 7.34 & 27.10 & 5.20 & \underline{44.72} & 7.06 & \underline{24.65} & 6.80 \\
& ProCap~{$_{\text{Vicuna-1.5-7B}}$ (ours)} & \underline{26.20} & 22.08 & \underline{86.26} & 14.88 & \underline{58.70} & 8.84 & 43.46 & 7.18 & \underline{27.32} & 5.30 & 43.94 & 7.10 & 24.44 & 6.60 \\
& Qwen3-VL-8B-Instruct~{$_{\text{RGBP}}$~\cite{bai2025qwen3vltechnicalreport}} & \textbf{38.18} & \textbf{28.90} & \textbf{136.60} & \textbf{22.24} & \textbf{104.50} & \textbf{13.54} & \textbf{113.76} & \textbf{14.36} & \textbf{103.94} & \textbf{13.30} & \textbf{108.66} & \textbf{13.76} & \textbf{85.82} & \textbf{17.02} \\
\bottomrule[1pt]
\end{tabular}
}
\end{table*}

\subsection{Results on RGBP benchmarks}
The RGBP evaluation set consists of 60 seen physical scenes (with novel/unseen projections) and 5 entirely unseen scenes. This setup tests both content recognition and environmental generalization.

\subsubsection{Performance on seen scenes}
As shown in~\cref{tab:results_60_seen_scenes}, ProCap variants significantly outperform off-the-shelf baselines in the \textbf{scene captioning task}. ProCap~{$_{\text{TinyLlama-1.1B}}$} achieves the highest CIDEr scores on COCO (70.27) and WHOOPS! (69.95), while ProCap~{$_{\text{OPT-2.7B}}$} provides the most accurate semantic structures, leading in SPICE scores ($\sim$24.0) across all subsets. These results show ProCap's capacity for robust scene understanding despite the presence of projected content.

In the \textbf{projection captioning task}, the fine-tuned Qwen3-VL-8B-Instruct~{$_{\text{RGBP}}$} achieves a dominant CIDEr of 127.58 on COCO, a nearly 11$\times$ improvement over the base Qwen3-VL-8B-Instruct model. The failure of baseline models (max CIDEr of 11.56) highlights the severity of virtual-physical ambiguity in standard VLM architectures. Among the ProCap variants, ProCap~{$_{\text{Vicuna-1.5-7B}}$} and ProCap~{$_{\text{OpenLLaMA-3B}}$} perform best, reaching CIDEr scores of 78.99 and 76.98 on COCO, respectively. ProCap with Qwen3-VL-8B-Instruct backbone was not tested due to time constraint\protect\footnotemark[1].

\subsubsection{Generalization to unseen scenes}
\cref{tab:results_5_unseen_scenes} shows performance on five scenes excluded from training. In scene captioning, ProCap series maintain a CIDEr advantage (\eg, ProCap~{$_{\text{OPT-2.7B}}$} at 5.80 vs. Qwen3-VL-8B at 3.68 on COCO). However, the base Qwen3-VL-8B model retains superior SPICE scores (12.26), likely due to its extensive pre-training on general scene-graph relationships.

For \textbf{projection captioning}, the advantage of our RGBP dataset remains clear. Qwen3-VL-8B-Instruct~{$_{\text{RGBP}}$} maintains a CIDEr of 136.60 on COCO and over 100 on nocaps variants. ProCap~{$_{\text{Vicuna-1.5-7B}}$} also shows strong generalization, significantly outperforming all non-fine-tuned baselines. This confirms that the region-aware retrieval remains effective even when the projected content  is unknown to the model.

\cref{fig:rgbp_qualitative_results} shows the descriptive capabilities of ProCap~{$_{\text{OpenLLaMA-3B}}$} compared to FastVLM-7B. In complex SAR scenes, FastVLM frequently suffers from virtual-physical ambiguity where it describes projected surfer, ducklings, cat, clock as physical ones (the first, second, fifth and sixth columns of \cref{fig:rgbp_qualitative_results}). In contrast, ProCap correctly identifies the projected content, even under geometric and photometric distortions. 

Our experiments show that projection captioning exhibits strong adaptability from seen to unseen scenes, yet a noticeable performance gap remains in scene captioning. This is likely due to the inherent challenges of SAR scenes, such as complex lighting and material properties, and the limited scale of real-world training data, despite capturing 60 scenes including 50 captions per scene, this is still small compared to over 118, 000 projection images and multiple captions per image in COCO~\cite{lin2014coco}. 

Our model also exhibits a distinct metric profile: higher CIDEr but lower SPICE scores compared to recently developed VLMs such as Qwen3-VL~\cite{bai2025qwen3vltechnicalreport}. This reflects our region-aware retrieval module enables accurate identification of domain-specific objects (\eg, LVIS categories), even under severe geometric and photometric distortions where zero-shot models tend to hallucinate, leading to strong CIDEr performance. In contrast, the lower SPICE scores indicate limited ability to model complex semantic relations, which is constrained by the capacity of the earlier LLMs. 
\begin{table}[!tbh]
    \centering
    \renewcommand{\arraystretch}{1.3}
    \caption{Ablation study of projection segmentation module on \textbf{the 5 unseen scenes with unseen projected content}. CIDEr (C), SPICE (S) are reported. The best results are in \textbf{bold}.}
    \label{tab:results_ablation_mask}

    \resizebox{\linewidth}{!}{
    \begin{tabular}{cl rr rr rr}
    \toprule[1pt]
    \multicolumn{1}{c}{\multirow{3}{*}{\textbf{Task}}} & \multirow{3}{*}{\textbf{Method}} & \multicolumn{2}{c}{\textbf{COCO}} & \multicolumn{2}{c}{\textbf{nocaps val}} & \multicolumn{2}{c}{\textbf{WHOOPS!}} \\
    \multicolumn{1}{c}{} & & \multicolumn{2}{c}{Test} & \multicolumn{2}{c}{Overall} & \multicolumn{2}{c}{Test} \\
    \cmidrule(lr){3-4} \cmidrule(lr){5-6} \cmidrule(lr){7-8} 
    \multicolumn{1}{c}{} & & \multicolumn{1}{c}{C $\uparrow$} & \multicolumn{1}{c}{S $\uparrow$} & \multicolumn{1}{c}{C $\uparrow$} & \multicolumn{1}{c}{S $\uparrow$} & \multicolumn{1}{c}{C $\uparrow$} & \multicolumn{1}{c}{S $\uparrow$} \\
    \midrule
    \addlinespace[0.5em]
    \multirow{1}{*}{\rotatebox[origin=c]{90}{Scene}} & 
    ProCap $_{\text{TinyLlama-1.1B}}$ & \textbf{5.02} & 5.53 & \textbf{4.06} & \textbf{8.34} & \textbf{7.48} & 5.15 \\
    & ProCap $_{\text{TinyLlama-1.1B w/o mask}}$ & 2.32 & \textbf{6.36} & 1.78 & 6.50 & 6.85 & \textbf{7.00} \\
   \midrule
    \multirow{1}{*}{\rotatebox[origin=c]{90}{Proj.}} & 
    ProCap $_{\text{TinyLlama-1.1B}}$ & 46.90 & \textbf{12.60} & \textbf{32.36} & 5.10 & 15.23 & 4.57 \\
    & ProCap $_{\text{TinyLlama-1.1B w/o mask}}$ & \textbf{56.50} & 10.68 & 25.58 & \textbf{5.20} & \textbf{19.70} & \textbf{5.98} \\
    \addlinespace[0.3em]
    \bottomrule[1pt]
    \end{tabular}
    }
\end{table}

\subsection{Ablation studies}
We conduct a series of ablation experiments to validate the individual contributions of the projection segmentation module, the region-aware retrieval mechanism, and the dual-captioning strategy.

\subsubsection{Effectiveness of the projection segmentation module}

We investigate the contribution of the projection segmentation module by comparing the framework with  variants that are without the refinement module $\mathcal{U}$ (w/o refinement) and without mask pooling (w/o mask). 

For the \textbf{mask pooling} mechanism, the results in \cref{tab:results_ablation_mask} show a critical trade-off. In the scene-captioning task, the inclusion of the mask is clearly superior for CIDEr scores. Interestingly, the SPICE score is often higher for the w/o mask variant in this task (e.g., 6.36 vs. 5.53 on COCO). This suggests that while the model loses descriptive detail (CIDEr), it may pick up broader semantic categories from the global image, though it fails to organize them into a precise caption. 

In the projection-captioning task, we see a counter-intuitive trend where the w/o mask variant sometimes achieves higher CIDEr scores (e.g., 56.50 on COCO and 19.70 on WHOOPS!). This may be caused by our coarse masking strategy, which may clip some important peripheral boundaries of the projection region. However, for a reliable SAR application, we must accurately describe the physical scene and projected content. Therefore, we choose the w/ mask version to balance the dual-captioning objectives.

For the \textbf{feature refinement} module, removing the deconvolutional upsampling and refinement layers $\mathcal{U}$, and directly inputting the coarse features to the segmentation and mask pooling modules (w/o refinement) leads to a significant degradation in projection detail (as shown in \cref{tab:results_ablation_module}), with CIDEr on COCO falling from 66.03 to 52.55. 

While SPICE scores remain stable (indicating basic category recognition), the drop in CIDEr suggests that refined features are vital for the quality of sentence structure and precise visual-textual alignment. The role of mask pooling is further evaluated by comparing ProCap with a variant trained without masks (w/o mask). 

\begin{table*}[t]
    \centering
    \renewcommand{\arraystretch}{1.2} 
    \caption{Ablation study of the feature refinement module and dual-captioning strategy on \textbf{the 5 unseen scenes with unseen projected content}. BLEU@4 (B@4), METEOR (M), CIDEr (C), SPICE (S) are reported. The best results are in \textbf{bold}.}
    \label{tab:results_ablation_module}

\resizebox{\linewidth}{!}{
\begin{tabular}{cl rrrr rrrrrrrr rr}
\toprule[1pt]
\multirow{3}{*}{\textbf{Task}} & \multirow{3}{*}{\textbf{Method}}
& \multicolumn{4}{c}{\textbf{COCO}} & \multicolumn{8}{c}{\textbf{nocaps val}} & \multicolumn{2}{c}{\textbf{WHOOPS!}} \\
& & \multicolumn{4}{c}{Test} & \multicolumn{2}{c}{In-domain} & \multicolumn{2}{c}{Near-domain} & \multicolumn{2}{c}{Out-domain} & \multicolumn{2}{c}{Overall} & \multicolumn{2}{c}{Test} \\
\cmidrule(lr){3-6} \cmidrule(lr){7-14} \cmidrule(lr){15-16}
& & \multicolumn{1}{c}{B@4 $\uparrow$} & \multicolumn{1}{c}{M $\uparrow$} & \multicolumn{1}{c}{C $\uparrow$} & \multicolumn{1}{c}{S $\uparrow$} & \multicolumn{1}{c}{C $\uparrow$} & \multicolumn{1}{c}{S $\uparrow$} & \multicolumn{1}{c}{C $\uparrow$} & \multicolumn{1}{c}{S $\uparrow$} & \multicolumn{1}{c}{C $\uparrow$} & \multicolumn{1}{c}{S $\uparrow$} & \multicolumn{1}{c}{C $\uparrow$} & \multicolumn{1}{c}{S $\uparrow$} & \multicolumn{1}{c}{C $\uparrow$} & \multicolumn{1}{c}{S $\uparrow$} \\
\midrule
\multirow{2}{*}[-1em]{%
  \raisebox{0.4ex}[0pt][0pt]{%
    \rotatebox[origin=c]{90}{%
      \shortstack[c]{Scene}%
    }%
  }%
}
& ProCap~{$_{\text{OpenLLaMA-3B}}$} & 1.50 & \textbf{11.73} & \textbf{3.78} & 7.77 & \textbf{4.88} & \textbf{7.16} & \textbf{4.96} & \textbf{7.28} & \textbf{5.56} & \textbf{7.14} & \textbf{4.02} & \textbf{7.10} & \textbf{5.73} &\textbf{ 6.60} \\
& ProCap~{$_{\text{OpenLLaMA-3B, w/o refinement}}$} & 0.25 & 10.60 & 3.68 & \textbf{7.78} & 4.56 & 6.10 & 4.34 & 6.28 & 4.62 & 6.46 & 3.60 & 6.28 & 5.52 & 5.73 \\
& ProCap~{$_{\text{OpenLLaMA-3B, w/o projection Q-Former}}$} & \textbf{1.88} & 10.60 & 3.80 & 6.67 & 4.74 & 5.88 & 4.22 & 5.26 & 4.12 & 5.94 & 3.42 & 5.60 & 6.18 & 5.42 \\
\midrule
\multirow{2}{*}[-0.8em]{%
  \raisebox{0.4ex}[0pt][0pt]{%
    \rotatebox[origin=c]{90}{%
      \shortstack[c]{Projection}%
    }%
  }%
}
& ProCap~{$_{\text{OpenLLaMA-3B}}$} & 19.90 & 21.22 & 66.03 & 14.90 & 58.22 & 8.66 & \textbf{48.00} & 7.34 & \textbf{27.10} & 5.20 & \textbf{44.72} & 7.06 & \textbf{24.65} & 6.80 \\
& ProCap~{$_{\text{OpenLLaMA-3B, w/o refinement}}$} & 9.90 & 19.90 & 52.55 & 15.60 & 47.02 & \textbf{9.00} & 38.30 & \textbf{7.44} & 24.42 & \textbf{5.38} & 37.92 & \textbf{7.26} & 15.22 & \textbf{7.75} \\
& ProCap~{$_{\text{OpenLLaMA-3B, w/o scene Q-Former}}$} & \textbf{27.02} & \textbf{23.72} & \textbf{67.38} & \textbf{16.63} & \textbf{60.20} & 8.70 & 46.66 & 7.22 & 23.54 & 4.24 & 44.22 & 6.76 & 22.18 & 6.57 \\

\bottomrule[1pt]
\end{tabular}
}
\end{table*}

\subsubsection{Effectiveness of the region-aware retrieval}
To evaluate the region-aware retrieval mechanism's ability to resolve virtual-physical ambiguity, we compare ProCap~{$_{\text{Vicuna-1.5-7B}}$} with a null variant (ProCap $_{\text{Vicuna-1.5-7B, w/o RAR.}}$) that is without the retrieved semantic context (\cref{tab:results_ablation_augment}). In the null variant, semantic context retrieval is performed, but object names $\boldsymbol{N}$ are replaced with null tokens.

As shown in \cref{tab:results_ablation_augment}, the inclusion of semantic context retrieved from the external knowledge base yields substantial gains in projection captioning. On the COCO subset, the CIDEr score increases from 67.98 to 86.26. A similar trend is observed in the nocaps overall benchmark, where CIDEr and SPICE improve by 7.66 and 1.62, respectively. These results confirm that semantic context is  essential to address virtual-physical ambiguity. Without high-fidelity semantic context, the VLM struggles to interpret projected content degraded by geometric and photometric distortions.

\begin{table}[!tbh]
    \centering
    \renewcommand{\arraystretch}{1.2} 
    \caption{Ablation study of region-aware retrieval (RAR) on \textbf{the 5 unseen scenes with unseen projected content}. CIDEr (C), SPICE (S) are reported. The best results are in \textbf{bold}.}
    \label{tab:results_ablation_augment}

\resizebox{\linewidth}{!}{
\begin{tabular}{l rr rr rr}
\toprule[1pt]
\multirow{3}{*}{\textbf{Method}}
& \multicolumn{2}{c}{\textbf{COCO}} & \multicolumn{2}{c}{\textbf{nocaps val}} & \multicolumn{2}{c}{\textbf{WHOOPS!}} \\
& \multicolumn{2}{c}{Test} & \multicolumn{2}{c}{Overall} & \multicolumn{2}{c}{Test} \\
\cmidrule(lr){2-3} \cmidrule(lr){4-5} \cmidrule(lr){6-7}
& \multicolumn{1}{c}{C $\uparrow$} & \multicolumn{1}{c}{S $\uparrow$} & \multicolumn{1}{c}{C $\uparrow$} & \multicolumn{1}{c}{S $\uparrow$} & \multicolumn{1}{c}{C $\uparrow$} & \multicolumn{1}{c}{S $\uparrow$} \\
\midrule
ProCap $_{\text{Vicuna-1.5-7B}}$ & \textbf{86.26} & \textbf{14.88} & \textbf{43.94} & \textbf{7.10} & \textbf{24.44} & \textbf{6.60} \\
ProCap $_{\text{Vicuna-1.5-7B, w/o RAR}}$ & 67.98 & 11.64 & 36.28 & 5.48 & 20.48 & 5.50 \\

\bottomrule[1pt]
\end{tabular}
}
\end{table}

\subsubsection{Effectiveness of the dual-captioning strategy}
We compare ProCap against specialist models trained only for a single captioning task. As shown in \cref{tab:results_ablation_module}, a scene-only specialist (w/o projection Q-Former) is clearly worse than a ProCap with dual-captioning capability. A projection-only specialist (w/o scene Q-Former) achieves a marginal CIDEr gain (67.38 vs. 66.03 CIDEr on COCO) but entirely loses the ability to describe the physical scene. ProCap achieves competitive performance with specialists while maintaining dual-task capability, proving that our framework effectively decouples the feature space without requiring separate task-specific backbones.

\section{Discussion}\label{sec:discussion}
\subsection{Applications}
The implications of ProCap and the RGBP dataset extend beyond decoupled captioning, offering a roadmap for integrating SAR into large-scale multimodal systems.

\subsubsection{ProCap as specialized module in MoE architectures.}
The development of VLMs is increasingly defined by the Mixture-of-Experts (MoE)~\cite{li2023multimodalfoundationmodelsspecialists} architecture. In this paradigm, a lightweight gating network (router), dynamically allocates tasks to specialized sub-networks (expert). While generalist VLMs excel at natural scene parsing, they remain susceptible to the virtual-physical ambiguity inherent in SAR. 

We position ProCap as a domain-specific expert within this MoE ecosystem. By decoupling physical and virtual layers, ProCap provides the precise spatial-semantic understanding required for agent-based SAR applications. Integrating ProCap as a specialized module allows a high-level router to bypass generalist reasoning when SAR elements are detected, ensuring architectural robustness and reducing the hallucination rates typically observed in monolithic backbones.

\subsubsection{SAR scene synthesis via natural language instructions}
Beyond serving as a benchmark for evaluating visual understanding, the RGBP dataset also enables a novel generative perspective on SAR scenes. Image captioning and image generation form a natural duality: captioning abstracts a visual scene into a compact semantic representation, while generation instantiates such semantics back into a concrete visual realization. This paradigm has driven significant breakthroughs in text-to-image generation, with large-scale image caption datasets such as COCO~\cite{lin2014coco} laying the foundation for modern models like Stable Diffusion~\cite{Rombach_2022_CVPR} and DALL-E~\cite{pmlr-v139-ramesh21a}. 

By providing factorized, dual-caption annotations, RGBP enables the training of caption-conditioned generative models for SAR. Unlike standard text-to-image datasets, our decoupled ground truth allows for independent control over the physical scene and the projected content. This is critical for synthesizing high-fidelity SAR training data, performing SAR simulation/relighting, and developing interactive design tools where users can manipulate projected content via natural language instructions~\cite{Deng2025LAPIG}.

\subsection{Limitations and future work}
Despite the effectiveness and robustness of the RGBP benchmark dataset and ProCap framework, several limitations remain. 
\subsubsection{Limtations of RGBP dataset}
\vspace{0.5em}\noindent\textbf{Simple geometries.} The current RGBP dataset focuses on fundamental projection scenarios where content is localized within rectangular boundaries and projected onto predominantly planar or mildly curved surfaces. Real-world SAR and projection mapping applications often involve complex, non-rigid, and even dynamic surfaces that not yet fully captured in our data. In such practical settings, projected content often blends near-seamlessly with the physical scenes, enabling a significantly higher degree of virtual-physical ambiguity than that observed in our current RGBP dataset. 

\vspace{0.5em}\noindent\textbf{Simple light transport.} Our dataset currently models the relationship between projected content and physical scenes as a layered overlay. While effective for decoupling, it does not fully simulate the complex light transport effects, such as ambient occlusion, specular reflections on non-Lambertian surfaces, and color bleeding, inherent in high-fidelity projection mapping or complex SAR environments.

Future work will focus on transitioning from simple layered overlays to spatially coherent and immersive SAR environment with dense dual-captioning annotations. This will enable the model to move beyond region-based recognition toward understanding the volumetric interaction between light and 3D physical structures.

\subsubsection{Limtations of ProCap framework}
\vspace{0.5em}\noindent\textbf{Sensitivity to the projection segmentation.}
ProCap's performance is coupled with the accuracy of the initial projection segmentation module. In cases involving highly specular or transparent surfaces, irregular boundary geometries, or low-contrast integration into the physical scene, segmentation inaccuracies can propagate to the downstream tasks. This  remains a bottleneck for achieving high-fidelity dual-captioning.

\vspace{0.5em}\noindent\textbf{Sensitivity to the external knowledge base scale.}
While our region-aware retrieval mechanism enhances descriptive detail, it is constrained by the vocabulary and concepts present in the LVIS-based knowledge base~\cite{li2024evcap,gupta2019lvis}. The retrieval of highly specialized, out-of-distribution (OOD) objects or novel domain-specific concepts remains challenging, particularly when the projected content is heavily degraded by environment.

Due to the iterative nature of VLM development, our current implementation utilizes established LLM backbones (e.g., Vicuna-1.5, OpenLLaMA). While our framework demonstrates the effectiveness of the architecture itself, we recognize that the recent emergence of more advanced backbones, such as the Qwen3-VL series (released after the submission deadline of this work), offers potential for even greater expressive power. The modular design of ProCap, however, ensures that it can be seamlessly updated with these state-of-the-art base models.

\section{Conclusion}\label{sec:conclusion}
This paper proposes ProCap framework to addresses the virtual-physical ambiguity and perceptual degradation  issues in SAR. By integrating an automated projection segmentation module with a region-aware semantic retrieval mechanism, ProCap produces high-fidelity dual captions for physical scene and projected content.  

To facilitate this research, we presented RGBP, the first large-scale SAR semantic benchmark featuring 180,000+ decoupled annotations. Evaluations under our dual-captioning protocol show advantages of ProCap over baselines, establishing a robust foundation for autonomous, context-aware SAR agents. This work is expected to facilitate transition for SAR from low-level geometric calibration toward high-level multimodal reasoning in complex environments.
\acknowledgments{
We thank the anonymous reviewers for their valuable and inspiring comments and suggestions.
}
\bibliographystyle{abbrv-doi}

\bibliography{ref}

@inproceedings{iwai2006limpiddesk,
  author    = {Iwai, Daisuke and Sato, Kosuke},
  title     = {Limpid desk: see-through access to disorderly desktop in projection-based mixed reality},
  year      = {2006},
  publisher = {ACM},
  booktitle = {Proceedings of the ACM Symposium on Virtual Reality Software and Technology},
  pages     = {112–115},
  numpages  = {4}
}

@article{bermano2017makeuplamps,
  author    = {Bermano, Amit H. and Billeter, Markus and Iwai, Daisuke and Grundh\"{o}fer, Anselm},
  title     = {{Makeup Lamps}: Live Augmentation of Human Faces via Projection},
  year      = {2017},
  publisher = {The Eurographs Association \& John Wiley \& Sons, Ltd.},
  volume    = {36},
  number    = {2},
  issn      = {0167-7055},
  journal   = {Computer Graphics Forum},
  pages     = {311–323},
  numpages  = {13}
}

@article{Kagami2019AnimatedStickies,
  author   = {Kagami, Shingo and Hashimoto, Koichi},
  journal  = {IEEE TVCG},
  title    = {{Animated Stickies}: Fast Video Projection Mapping onto a Markerless Plane through a Direct Closed-Loop Alignment},
  year     = {2019},
  volume   = {25},
  number   = {11},
  pages    = {3094-3104}
}

@inproceedings{Peng2020high,
  author    = {Peng, Hao-Lun and Watanabe, Yoshihiro},
  title     = {High-Speed Human Arm Projection Mapping with Skin Deformation},
  year      = {2020},
  publisher = {ACM},
  booktitle = {SIGGRAPH Asia},
  articleno = {13},
  numpages  = {2}
}

@inproceedings{Asahina2021realistic,
  author    = {Asahina, Ray and Nomoto, Takashi and Yoshida, Takatoshi and Watanabe, Yoshihiro},
  booktitle = {IEEE VR},
  title     = {Realistic 3D Swept-Volume Display with Hidden-Surface Removal Using Physical Materials},
  year      = {2021},
  volume    = {},
  number    = {},
  pages     = {113-121}
}

@article{erel2023neuralprojectionmapping,
  author   = {Erel, Yotam and Iwai, Daisuke and Bermano, Amit H.},
  journal  = {IEEE TVCG},
  title    = {Neural Projection Mapping Using Reflectance Fields},
  year     = {2023},
  volume   = {29},
  number   = {11},
  pages    = {4339-4349}
}

@inproceedings{erel2024casperdpm,
  author    = {Erel, Yotam and Kozlovsky-Mordenfeld, Or and Iwai, Daisuke and Sato, Kosuke and Bermano, Amit H.},
  title     = {{Casper DPM}: Cascaded Perceptual Dynamic Projection Mapping onto Hands},
  year      = {2024},
  publisher = {ACM},
  booktitle = {SIGGRAPH Asia},
  articleno = {137},
  numpages  = {10}
}

@inproceedings{dong2023Calibration,
  author    = {Dong, Xin and Ling, Haibin and Huang, Bingyao},
  booktitle = {IEEE International Symposium on Mixed and Augmented Reality},
  title     = {Adaptive Color Structured Light for Calibration and Shape Reconstruction},
  year      = {2023},
  volume    = {},
  number    = {},
  pages     = {1240-1249}
}

@article{Raskar2003iLamps,
  author     = {Raskar, Ramesh and van Baar, Jeroen and Beardsley, Paul and Willwacher, Thomas and Rao, Srinivas and Forlines, Clifton},
  title      = {{iLamps}: geometrically aware and self-configuring projectors},
  year       = {2003},
  volume     = {22},
  number     = {3},
  issn       = {0730-0301},
  journal    = {ACM TOG},
  pages      = {809–818},
  numpages   = {10}
}

@article{grundhofer2015robust,
  title     = {Robust, error-tolerant photometric projector compensation},
  author    = {Grundh{\"o}fer, Anselm and Iwai, Daisuke},
  journal   = {IEEE TIP},
  volume    = {24},
  number    = {12},
  pages     = {5086--5099},
  year      = {2015},
  publisher = {IEEE}
}

@article{huang2021flexible,
    title={A Fast and Flexible Projector-Camera Calibration System}, 
    author={Huang, Bingyao and Tang, Ying and Ozdemir, Samed and Ling, Haibin},
    journal={IEEE Transactions on Automation Science and Engineering}, 
    year={2021},
}

@INPROCEEDINGS{kaminokado2019augment,
  author={Kaminokado, Takumi and Iwai, Daisuke and Sato, Kosuke},
  booktitle={IEEE International Symposium on Mixed and Augmented Reality}, 
  title={Augmented Environment Mapping for Appearance Editing of Glossy Surfaces}, 
  year={2019},
  volume={},
  number={},
  pages={55-65}}

@inproceedings{Nomoto2020dpm,
author = {Nomoto, Takashi and Koishihara, Ryo and Watanabe, Yoshihiro},
title = {Realistic Dynamic Projection Mapping Using Real-Time Ray Tracing},
year = {2020},
publisher = {ACM},
booktitle = {SIGGRAPH},
articleno = {13},
numpages = {2}
}

@inproceedings{Nomoto2020multiproj,
author = {Nomoto, Takashi and Li, Wanlong and Peng, Hao-Lun and Watanabe, Yoshihiro},
title = {Dynamic Projection Mapping with Networked Multi-projectors Based on Pixel-parallel Intensity Control},
year = {2020},
publisher = {ACM},
booktitle = {SIGGRAPH Asia},
articleno = {11},
numpages = {2},
}

@ARTICLE{Nomoto2022dpm,
  author={Nomoto, Takashi and Li, Wanlong and Peng, Hao-Lun and Watanabe, Yoshihiro},
  journal={IEEE TVCG}, 
  title={Dynamic Multi-projection Mapping Based on Parallel Intensity Control}, 
  year={2022},
  volume={28},
  number={5},
  pages={2125-2134}}

@ARTICLE{peng2025fp,
  author={Peng, Hao-Lun and Sato, Kengo and Nakagawa, Soran and Watanabe, Yoshihiro},
  journal={IEEE TVCG}, 
  title={Perceptually-Aligned Dynamic Facial Projection Mapping by High-Speed Face-Tracking Method and Lens-Shift Co-Axial Setup}, 
  year={2025},
  volume={31},
  number={10},
  pages={6824-6838}}

@article{huang2017radiometriccompensation,
  author  = {Huang, Tai-Hsiang and Wang, Ting-Chun and Chen, Homer H.},
  journal = {IEEE TIP},
  title   = {Radiometric Compensation of Images Projected on Non-White Surfaces by Exploiting Chromatic Adaptation and Perceptual Anchoring},
  year    = {2017},
  volume  = {26},
  number  = {1},
  pages   = {147-159}
}

@inproceedings{huang2019compennet,
  author    = {Huang, Bingyao and Ling, Haibin},
  title     = {End-To-End Projector Photometric Compensation},
  booktitle = {CVPR},
  year      = {2019},
  volume    = {},
  number    = {},
  pages     = {6803-6812},
}

@article{huang2022CompenNeSt++,
  title   = {End-to-end Full Projector Compensation},
  author  = {Bingyao Huang and Tao Sun and Haibin Ling},
  year    = {2022},
  journal = {IEEE TPAMI}
}

@article{huang2021deprocams,
  author   = {Huang, Bingyao and Ling, Haibin},
  journal  = {IEEE TVCG},
  title    = {{DeProCams}: Simultaneous Relighting, Compensation and Shape Reconstruction for Projector-Camera Systems},
  year     = {2021},
  volume   = {27},
  number   = {5},
  pages    = {2725-2735}
}

@inproceedings{luo2021staypositive,
  author    = {Luo, Katie and Yang, Guandao and Xian, Wenqi and Haraldsson, Harald and Hariharan, Bharath and Belongie, Serge},
  title     = {{Stay Positive}: Non-Negative Image Synthesis for Augmented Reality},
  booktitle = {CVPR},
  year      = {2021},
  pages     = {10050-10060}
}

@inproceedings{wang2023CompenHR,
  author    = {Wang, Yuxi and Ling, Haibin and Huang, Bingyao},
  booktitle = {IEEE VR},
  title     = {{CompenHR}: Efficient Full Compensation for High-resolution Projector},
  year      = {2023},
  volume    = {},
  number    = {},
  pages     = {135-145}
}

@article{wang2024vicomp,
  author  = {Wang, Yuxi and Ling, Haibin and Huang, Bingyao},
  journal = {IEEE TVCG},
  title   = {{ViComp}: Video Compensation for Projector-Camera Systems},
  year    = {2024},
  volume  = {30},
  number  = {5},
  pages   = {2347-2356}
}

@ARTICLE{Kageyama2022deblur,
  author={Kageyama, Yuta and Iwai, Daisuke and Sato, Kosuke},
  journal={IEEE TVCG}, 
  title={Online Projector Deblurring Using a Convolutional Neural Network}, 
  year={2022},
  volume={28},
  number={5},
  pages={2223-2233}}

@article{Kusuyama2024deblur,
  author  = {Kusuyama, Hiroki and Kageyama, Yuta and Iwai, Daisuke and Sato, Kosuke},
  journal = {IEEE TVCG},
  title   = {A Multi-aperture Coaxial Projector Balancing Shadow Suppression and Deblurring},
  year    = {2024},
  volume  = {},
  number  = {},
  pages   = {1-11}
}

@ARTICLE{Deng2025LAPIG,
  author  = {Deng, Yuchen and Ling, Haibin and Huang, Bingyao},
  journal = {IEEE TVCG},
  title   = {{LAPIG}: Language Guided Projector Image Generation with Surface Adaptation and Stylization},
  year    = {2025}
}

@ARTICLE{Li2025DPCS,
    author  = {Li, Jijiang and Deng, Qingyue and Ling, Haibin and Huang, Bingyao},
    journal = {IEEE TVCG},
    title   = {{DPCS}: Path Tracing-Based Differentiable Projector-Camera Systems},
    year    = {2025}
}

@article{yasui2024pm,
  author  = {Yasui, Masahiko and Iwataki, Ryota and Ishikawa, Masatoshi and Watanabe, Yoshihiro},
  journal = {IEEE TVCG},
  title   = {Projection Mapping with a Brightly Lit Surrounding Using a Mixed Light Field Approach},
  year    = {2024},
  volume  = {30},
  number  = {5},
  pages   = {2217-2227}
}

@INPROCEEDINGS{Bitton-Guetta2023WHOOPS,
  author={Bitton-Guetta, Nitzan and Bitton, Yonatan and Hessel, Jack and Schmidt, Ludwig and Elovici, Yuval and Stanovsky, Gabriel and Schwartz, Roy},
  booktitle={ICCV}, 
  title={{Breaking Common Sense: WHOOPS!} A Vision-and-Language Benchmark of Synthetic and Compositional Images}, 
  year={2023},
  volume={},
  number={},
  pages={2616-2627}
}

@inproceedings{lin2014coco,
  author    = {Lin, Tsung-Yi
               and Maire, Michael
               and Belongie, Serge
               and Hays, James
               and Perona, Pietro
               and Ramanan, Deva
               and Doll{\'a}r, Piotr
               and Zitnick, C. Lawrence},
  title     = {{Microsoft COCO}: Common Objects in Context},
  booktitle = {ECCV},
  year      = {2014},
  pages     = {740--755}
}

@INPROCEEDINGS{li2024evcap,
  author={Li, Jiaxuan and Vo, Duc Minh and Sugimoto, Akihiro and Nakayama, Hideki},
  booktitle={CVPR}, 
  title={{EVCap}: Retrieval-Augmented Image Captioning with External Visual-Name Memory for Open-World Comprehension}, 
  year={2024},
  volume={},
  number={},
  pages={13733-13742}}

@INPROCEEDINGS{Agrawal2019nocaps,
  author={Agrawal, Harsh and Desai, Karan and Wang, Yufei and Chen, Xinlei and Jain, Rishabh and Johnson, Mark and Batra, Dhruv and Parikh, Devi and Lee, Stefan and Anderson, Peter},
  booktitle={ICCV}, 
  title={nocaps: novel object captioning at scale}, 
  year={2019},
  volume={},
  number={},
  pages={8947-8956}
}

@InProceedings{radford2021clip,
  author={Radford, Alec and Kim, Jong Wook and Hallacy, Chris and Ramesh, Aditya and Goh, Gabriel and Agarwal, Sandhini and Sastry, Girish and Askell, Amanda and Mishkin, Pamela and Clark, Jack and Krueger, Gretchen and Sutskever, Ilya},
  booktitle={ICML},
  title={Learning Transferable Visual Models From Natural Language Supervision},
  year={2021},
  volume={139},
  pages={8748--8763}
}

@InProceedings{li2022blip,
  title={{BLIP}: Bootstrapping Language-Image Pre-training for Unified Vision-Language Understanding and Generation},
  author={Li, Junnan and Li, Dongxu and Xiong, Caiming and Hoi, Steven},
  booktitle={ICML},
  pages={12888--12900},
  year={2022},
  volume={162},
}

@inproceedings{gupta2019lvis,
  title={{LVIS}: A Dataset for Large Vocabulary Instance Segmentation},
  author={Gupta, Agrim and Dollar, Piotr and Girshick, Ross},
  booktitle={CVPR},
  year={2019}
}

@inproceedings{blip2Qformer,
author = {Li, Junnan and Li, Dongxu and Savarese, Silvio and Hoi, Steven},
title = {{BLIP-2}: bootstrapping language-image pre-training with frozen image encoders and large language models},
year = {2023},
booktitle = {ICML}
}

@ARTICLE{takeuchi2024projlight,
  author={Takeuchi, Masaki and Kusuyama, Hiroki and Iwai, Daisuke and Sato, Kosuke},
  journal={IEEE TVCG}, 
  title={Projection Mapping under Environmental Lighting by Replacing Room Lights with Heterogeneous Projectors}, 
  year={2024},
  volume={30},
  number={5}}

@InProceedings{fastvlm2025,
  author = {Pavan Kumar Anasosalu Vasu and Fartash Faghri and Chun-Liang Li and Cem Koc and Nate True and Albert Antony and Gokul Santhanam and James Gabriel and Peter Grasch and Oncel Tuzel and Hadi Pouransari},
  title = {{FastVLM}: Efficient Vision Encoding for Vision Language Models},
  booktitle = {CVPR},
  year = {2025},
}

@misc{zhang2024tinyllama,
      title={{TinyLlama}: An Open-Source Small Language Model}, 
      author={Peiyuan Zhang and Guangtao Zeng and Tianduo Wang and Wei Lu},
      year={2024},
      eprint={2401.02385},
      archivePrefix={arXiv},
      primaryClass={cs.CL}
}

@misc{openlm2023openllama,
  author = {Geng, Xinyang and Liu, Hao},
  title = {{OpenLLaMA}: An Open Reproduction of {LLaMA}},
  year = {2023},
  url = {https://github.com/openlm-research/open_llama}
}

@article{touvron2023llama,
  title={Llama: Open and efficient foundation language models},
  author={Touvron, Hugo and Lavril, Thibaut and Izacard, Gautier and Martinet, Xavier and Lachaux, Marie-Anne and Lacroix, Timoth{\'e}e and Rozi{\`e}re, Baptiste and Goyal, Naman and Hambro, Eric and Azhar, Faisal and others},
  journal={arXiv preprint arXiv:2302.13971},
  year={2023}
}

@article{li2023multimodalfoundationmodelsspecialists,
    author = {Li, Chunyuan and Gan, Zhe and Yang, Zhengyuan and Yang, Jianwei and Li, Linjie and Wang, Lijuan and Gao, Jianfeng},
    title = {Multimodal Foundation Models: From Specialists to General-Purpose Assistants},
    journal = {Foundations and Trends in Computer Graphics and Vision},
    volume = {16},
    number = {1-2},
    pages = {1-214},
    year = {2024},
    issn = {1572-2740},
}

@article{DaisukeIWAI2024pjab.100.012,
  title={Projection mapping technologies: A review of current trends and future directions},
  author={Daisuke Iwai},
  journal={Proceedings of the Japan Academy. Series B, Physical and Biological Sciences},
  volume={100},
  number={3},
  pages={234-251},
  year={2024},
}

@InProceedings{jia2021scaling,
  title = 	 {Scaling Up Visual and Vision-Language Representation Learning With Noisy Text Supervision},
  author =       {Jia, Chao and Yang, Yinfei and Xia, Ye and Chen, Yi-Ting and Parekh, Zarana and Pham, Hieu and Le, Quoc and Sung, Yun-Hsuan and Li, Zhen and Duerig, Tom},
  booktitle = 	 {ICML},
  pages = 	 {4904--4916},
  year = 	 {2021},
  volume = 	 {139},
  series = 	 {Proceedings of Machine Learning Research},
  publisher =    {PMLR},
}

@INPROCEEDINGS{Wang2023Learning,
  author={Wang, Tzu-Jui Julius and Laaksonen, Jorma and Langer, Tomas and Arponen, Heikki and Bishop, Tom E.},
  booktitle={IEEE/CVF Winter Conference on Applications of Computer Vision (WACV)}, 
  title={{Learning by Hallucinating}: Vision-Language Pre-training with Weak Supervision}, 
  year={2023},
  volume={},
  number={},
  pages={1073-1083}}

@inproceedings{alayrac2022flamingo,
 author = {Alayrac, Jean-Baptiste and Donahue, Jeff and Luc, Pauline and Miech, Antoine and Barr, Iain and Hasson, Yana and Lenc, Karel and Mensch, Arthur and Millican, Katherine and Reynolds, Malcolm and Ring, Roman and Rutherford, Eliza and Cabi, Serkan and Han, Tengda and Gong, Zhitao and Samangooei, Sina and Monteiro, Marianne and Menick, Jacob L and Borgeaud, Sebastian and Brock, Andy and Nematzadeh, Aida and Sharifzadeh, Sahand and Bi\'{n}kowski, Miko\l aj and Barreira, Ricardo and Vinyals, Oriol and Zisserman, Andrew and Simonyan, Kar\'{e}n},
 booktitle = {NeurIPS},
 editor = {S. Koyejo and S. Mohamed and A. Agarwal and D. Belgrave and K. Cho and A. Oh},
 pages = {23716--23736},
 title = {Flamingo: a Visual Language Model for Few-Shot Learning},
 volume = {35},
 year = {2022}
}

@inproceedings{dai2023instructblip,
 author = {Dai, Wenliang and Li, Junnan and LI, DONGXU and Tiong, Anthony and Zhao, Junqi and Wang, Weisheng and Li, Boyang and Fung, Pascale N and Hoi, Steven},
 booktitle = {NeurIPS},
 editor = {A. Oh and T. Naumann and A. Globerson and K. Saenko and M. Hardt and S. Levine},
 pages = {49250--49267},
 title = {{InstructBLIP}: Towards General-purpose Vision-Language Models with Instruction Tuning},
 volume = {36},
 year = {2023}
}

@inproceedings{gururangan2020don,
    title = "Don{'}t Stop Pretraining: Adapt Language Models to Domains and Tasks",
    author = "Gururangan, Suchin and Marasovi{\'c}, Ana and Swayamdipta, Swabha and Lo, Kyle and Beltagy, Iz and Downey, Doug and Smith, Noah A.",
    booktitle = "Proceedings of the 58th Annual Meeting of the Association for Computational Linguistics",
    year = "2020",
    publisher = "Association for Computational Linguistics",
}

@InProceedings{houlsby2019parameter,
  title = 	 {Parameter-Efficient Transfer Learning for {NLP}},
  author =       {Houlsby, Neil and Giurgiu, Andrei and Jastrzebski, Stanislaw and Morrone, Bruna and De Laroussilhe, Quentin and Gesmundo, Andrea and Attariyan, Mona and Gelly, Sylvain},
  booktitle = 	 {ICML},
  pages = 	 {2790--2799},
  year = 	 {2019},
  volume = 	 {97},
  series = 	 {Proceedings of Machine Learning Research},
  publisher =    {PMLR},
}

@inproceedings{hu2022lora,
title={Lo{RA}: Low-Rank Adaptation of Large Language Models},
author={Edward J Hu and Yelong Shen and Phillip Wallis and Zeyuan Allen-Zhu and Yuanzhi Li and Shean Wang and Lu Wang and Weizhu Chen},
booktitle={ICLR},
year={2022},
}

@inproceedings{izacard2021leveraging,
    title = "Leveraging Passage Retrieval with Generative Models for Open Domain Question Answering",
    author = "Izacard, Gautier  and
      Grave, Edouard",
    booktitle = "Proceedings of the 16th Conference of the European Chapter of the Association for Computational Linguistics: Main Volume",
    year = "2021",
    publisher = "Association for Computational Linguistics",
    pages = "874--880",
}

@inproceedings{lewis2020retrieval,
 author = {Lewis, Patrick and Perez, Ethan and Piktus, Aleksandra and Petroni, Fabio and Karpukhin, Vladimir and Goyal, Naman and K\"{u}ttler, Heinrich and Lewis, Mike and Yih, Wen-tau and Rockt\"{a}schel, Tim and Riedel, Sebastian and Kiela, Douwe},
 booktitle = {NeurIPS},
 pages = {9459--9474},
 title = {Retrieval-Augmented Generation for Knowledge-Intensive {NLP} Tasks},
 volume = {33},
 year = {2020}
}

@InProceedings{Yasunaga2023retrieval,
  title = 	 {Retrieval-Augmented Multimodal Language Modeling},
  author =       {Yasunaga, Michihiro and Aghajanyan, Armen and Shi, Weijia and James, Richard and Leskovec, Jure and Liang, Percy and Lewis, Mike and Zettlemoyer, Luke and Yih, Wen-Tau},
  booktitle = 	 {ICML},
  pages = 	 {39755--39769},
  year = 	 {2023},
  volume = 	 {202},
  series = 	 {Proceedings of Machine Learning Research},
  publisher =    {PMLR},
}

@misc{bai2025qwen3vltechnicalreport,
      title={{Qwen3-VL} Technical Report}, 
      author={Shuai Bai and Yuxuan Cai and Ruizhe Chen and Keqin Chen and Xionghui Chen and Zesen Cheng and Lianghao Deng and Wei Ding and Chang Gao and Chunjiang Ge and Wenbin Ge and Zhifang Guo and Qidong Huang and Jie Huang and Fei Huang and Binyuan Hui and Shutong Jiang and Zhaohai Li and Mingsheng Li and Mei Li and Kaixin Li and Zicheng Lin and Junyang Lin and Xuejing Liu and Jiawei Liu and Chenglong Liu and Yang Liu and Dayiheng Liu and Shixuan Liu and Dunjie Lu and Ruilin Luo and Chenxu Lv and Rui Men and Lingchen Meng and Xuancheng Ren and Xingzhang Ren and Sibo Song and Yuchong Sun and Jun Tang and Jianhong Tu and Jianqiang Wan and Peng Wang and Pengfei Wang and Qiuyue Wang and Yuxuan Wang and Tianbao Xie and Yiheng Xu and Haiyang Xu and Jin Xu and Zhibo Yang and Mingkun Yang and Jianxin Yang and An Yang and Bowen Yu and Fei Zhang and Hang Zhang and Xi Zhang and Bo Zheng and Humen Zhong and Jingren Zhou and Fan Zhou and Jing Zhou and Yuanzhi Zhu and Ke Zhu},
      year={2025},
      eprint={2511.21631},
      archivePrefix={arXiv},
      primaryClass={cs.CV},
      url={https://arxiv.org/abs/2511.21631}, 
}

@misc{bai2025qwen25vltechnicalreport,
      title={{Qwen2.5-VL} Technical Report}, 
      author={Shuai Bai and Keqin Chen and Xuejing Liu and Jialin Wang and Wenbin Ge and Sibo Song and Kai Dang and Peng Wang and Shijie Wang and Jun Tang and Humen Zhong and Yuanzhi Zhu and Mingkun Yang and Zhaohai Li and Jianqiang Wan and Pengfei Wang and Wei Ding and Zheren Fu and Yiheng Xu and Jiabo Ye and Xi Zhang and Tianbao Xie and Zesen Cheng and Hang Zhang and Zhibo Yang and Haiyang Xu and Junyang Lin},
      year={2025},
      eprint={2502.13923},
      archivePrefix={arXiv},
      primaryClass={cs.CV},
      url={https://arxiv.org/abs/2502.13923}, 
}

@misc{zhang2022optopenpretrainedtransformer,
      title={{OPT}: Open Pre-trained Transformer Language Models}, 
      author={Susan Zhang and Stephen Roller and Naman Goyal and Mikel Artetxe and Moya Chen and Shuohui Chen and Christopher Dewan and Mona Diab and Xian Li and Xi Victoria Lin and Todor Mihaylov and Myle Ott and Sam Shleifer and Kurt Shuster and Daniel Simig and Punit Singh Koura and Anjali Sridhar and Tianlu Wang and Luke Zettlemoyer},
      year={2022},
      eprint={2205.01068},
      archivePrefix={arXiv},
      primaryClass={cs.CL},
      url={https://arxiv.org/abs/2205.01068}, 
}

@misc{zheng2023judgingllmasajudgemtbenchchatbot,
      title={Judging {LLM-as-a-Judge} with {MT-Bench} and {Chatbot Arena}}, 
      author={Lianmin Zheng and Wei-Lin Chiang and Ying Sheng and Siyuan Zhuang and Zhanghao Wu and Yonghao Zhuang and Zi Lin and Zhuohan Li and Dacheng Li and Eric P. Xing and Hao Zhang and Joseph E. Gonzalez and Ion Stoica},
      year={2023},
      eprint={2306.05685},
      archivePrefix={arXiv},
      primaryClass={cs.CL},
      url={https://arxiv.org/abs/2306.05685}, 
}

@misc{Qwen2-VL-Finetuning,
  author = {Yuwon Lee},
  title = {{Qwen2-VL-Finetune}},
  year = {2024},
  publisher = {GitHub},
  url = {https://github.com/2U1/Qwen2-VL-Finetune}
}

@InProceedings{Rombach_2022_CVPR,
    author    = {Rombach, Robin and Blattmann, Andreas and Lorenz, Dominik and Esser, Patrick and Ommer, Bj\"orn},
    title     = {High-Resolution Image Synthesis With Latent Diffusion Models},
    booktitle = {CVPR},
    year      = {2022},
    pages     = {10684-10695}
}

@InProceedings{pmlr-v139-ramesh21a,
  title = 	 {Zero-Shot Text-to-Image Generation},
  author =       {Ramesh, Aditya and Pavlov, Mikhail and Goh, Gabriel and Gray, Scott and Voss, Chelsea and Radford, Alec and Chen, Mark and Sutskever, Ilya},
  booktitle = 	 {ICMR},
  pages = 	 {8821--8831},
  year = 	 {2021},
  volume = 	 {139},
  series = 	 {Proceedings of Machine Learning Research},
  publisher =    {PMLR},
}

@book{Bimber2005sar,
author = {Bimber, Oliver and Raskar, Ramesh},
title = {Spatial Augmented Reality: Merging Real and Virtual Worlds},
year = {2005},
publisher = {A. K. Peters, Ltd.},
}

@misc{llama3modelcard,
title={Llama 3 Model Card},
author={AI@Meta},
year={2024},
url = {https://github.com/meta-llama/llama3/blob/main/MODEL_CARD.md}
}

@misc{zhu2025internvl3exploringadvancedtraining,
      title={{InternVL3}: Exploring Advanced Training and Test-Time Recipes for Open-Source Multimodal Models}, 
      author={Jinguo Zhu and Weiyun Wang and Zhe Chen and Zhaoyang Liu and Shenglong Ye and Lixin Gu and Hao Tian and Yuchen Duan and Weijie Su and Jie Shao and Zhangwei Gao and Erfei Cui and Xuehui Wang and Yue Cao and Yangzhou Liu and Xingguang Wei and Hongjie Zhang and Haomin Wang and Weiye Xu and Hao Li and Jiahao Wang and Nianchen Deng and Songze Li and Yinan He and Tan Jiang and Jiapeng Luo and Yi Wang and Conghui He and Botian Shi and Xingcheng Zhang and Wenqi Shao and Junjun He and Yingtong Xiong and Wenwen Qu and Peng Sun and Penglong Jiao and Han Lv and Lijun Wu and Kaipeng Zhang and Huipeng Deng and Jiaye Ge and Kai Chen and Limin Wang and Min Dou and Lewei Lu and Xizhou Zhu and Tong Lu and Dahua Lin and Yu Qiao and Jifeng Dai and Wenhai Wang},
      year={2025},
      eprint={2504.10479},
      archivePrefix={arXiv},
      primaryClass={cs.CV},
      url={https://arxiv.org/abs/2504.10479}, 
}

@inproceedings{Papineni2002bleu,
    title = "{B}leu: a Method for Automatic Evaluation of Machine Translation",
    author = "Papineni, Kishore  and
      Roukos, Salim  and
      Ward, Todd  and
      Zhu, Wei-Jing",
    booktitle = "Proceedings of the 40th Annual Meeting of the Association for Computational Linguistics",
    year = "2002",
    publisher = "Association for Computational Linguistics",
    pages = "311--318"
}

@inproceedings{banerjee2005meteor,
    title = "{METEOR}: An Automatic Metric for {MT} Evaluation with Improved Correlation with Human Judgments",
    author = "Banerjee, Satanjeev  and
      Lavie, Alon",
    booktitle = "Proceedings of the {ACL} Workshop on Intrinsic and Extrinsic Evaluation Measures for Machine Translation and/or Summarization",
    year = "2005",
    publisher = "Association for Computational Linguistics",
    pages = "65--72"
}

@InProceedings{Vedantam2015CIDEr,
author = {Vedantam, Ramakrishna and Lawrence Zitnick, C. and Parikh, Devi},
title = {{CIDEr}: Consensus-Based Image Description Evaluation},
booktitle = {CVPR},
year = {2015}
}

@inproceedings{spice2016,
  title     = {{SPICE}: Semantic Propositional Image Caption Evaluation},
  author    = {Peter Anderson and Basura Fernando and Mark Johnson and Stephen Gould},
  year      = {2016},
  booktitle = {ECCV}
}

@ARTICLE{xiu2025viddar,
  author={Xiu, Yanming and Scargill, Tim and Gorlatova, Maria},
  journal={IEEE TVCG}, 
  title={{ViDDAR}: Vision Language Model-Based Task-Detrimental Content Detection for Augmented Reality}, 
  year={2025},
  volume={31},
  number={5},
  pages={3194-3203},}

@INPROCEEDINGS{duan2025advancing,
  author={Duan, Lin and Xiu, Yanming and Gorlatova, Maria},
  booktitle={IEEE Conference on Virtual Reality and 3D User Interfaces Abstracts and Workshops (VRW)}, 
  title={Advancing the Understanding and Evaluation of {AR}-Generated Scenes: When Vision-Language Models Shine and Stumble}, 
  year={2025},
  volume={},
  number={},
  pages={156-161},}

@InProceedings{Urakawa2025neural,
    author    = {Urakawa, Yuki and Watanabe, Yoshihiro},
    title     = {Neural Inverse Rendering for High-Accuracy 3D Measurement of Moving Objects with Fewer Phase-Shifting Patterns},
    booktitle = {ICCV},
    year      = {2025},
    pages     = {27692-27701}
}

@ARTICLE{Deng2025GS-ProCams,
  author={Deng, Qingyue and Li, Jijiang and Ling, Haibin and Huang, Bingyao},
  journal={IEEE TVCG}, 
  title={{GS-ProCams}: Gaussian Splatting-Based Projector-Camera Systems}, 
  year={2025}
}

@InProceedings{iwai2025pm,
author="Iwai, Daisuke",
title="Projection Mapping Technologies for Seamless Spatial Augmentation",
booktitle="Proceedings of Laser Display and Lighting Conference",
year="2025",
publisher="Springer Nature Singapore",
pages="181--185",
}

@misc{sun2023EVA-CLIP,
  title={{EVA-CLIP}: Improved Training Techniques for CLIP at Scale},
  author={Sun, Quan and Fang, Yuxin and Wu, Ledell and Wang, Xinlong and Cao, Yue},
  year={2023},
  eprint={2303.15389},
  archivePrefix={arXiv},
}

\clearpage

\maketitlesupplementary

\appendix  

\setcounter{equation}{0}
\setcounter{figure}{0}
\setcounter{table}{0}
\setcounter{page}{1}

\section*{A. Introduction}
\begin{figure*}[t]
    \centering
    \includegraphics[width=\textwidth]{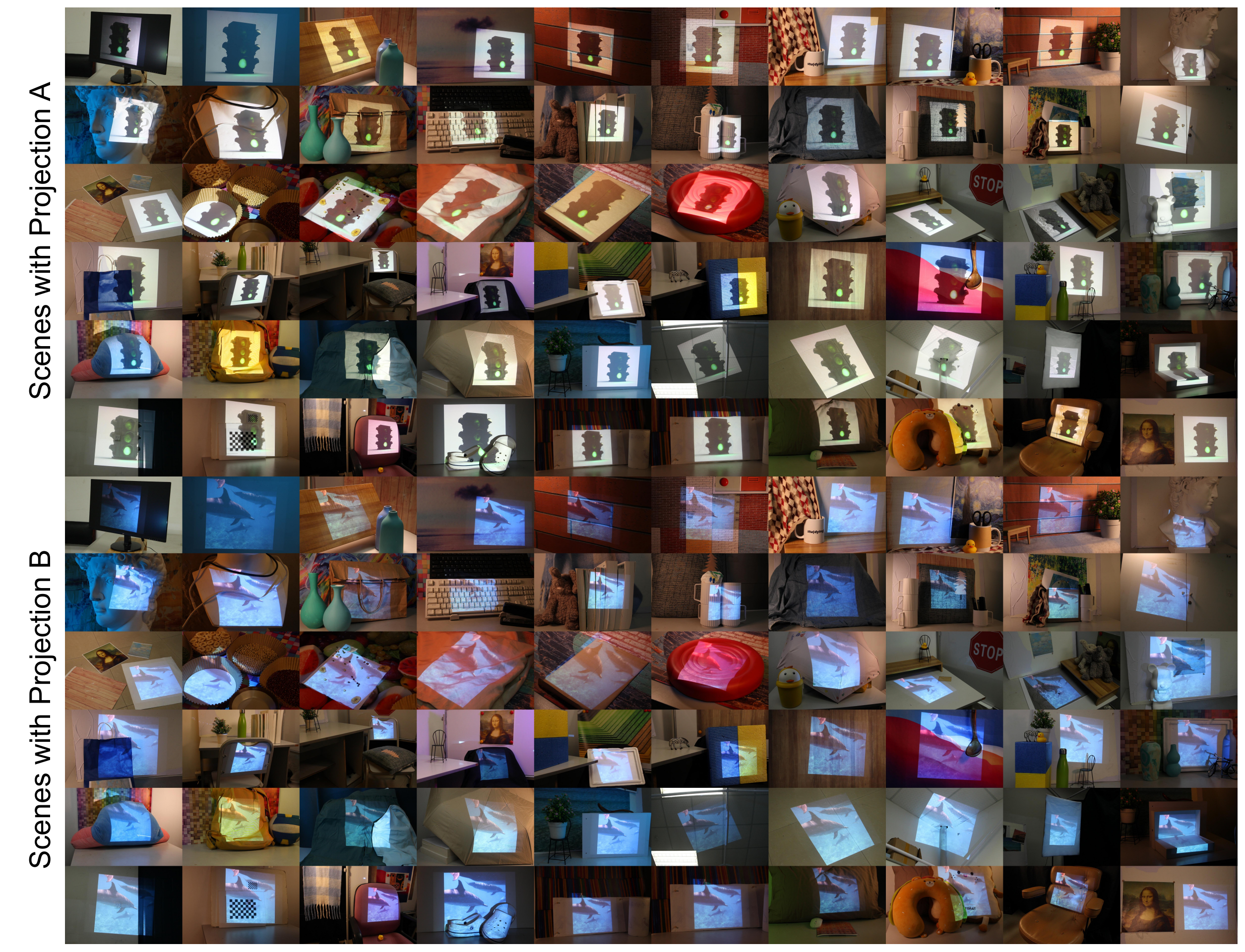}
    \caption{This figure shows the specific details of 60 seen scenes with projected content A (traffic lights) and projected content B (dolphins).}
    \label{fig:rgbp_samples}
\end{figure*}

This supplementary material provides extended experimental results and additional qualitative examples to complement main paper. We evaluate ProCap and competing baselines on all 60 seen scenes (More examples are shown in~\cref{fig:rgbp_samples}), 5 unseen scenes. Besides main paper, additional 5 unseen scenes are captured using an EPSON EB-C2050WN 3LCD projector (1600 × 1200) and a Nikon D3200 DSLR camera (1080 × 720), which differ from the hardware configurations used in the previous 65 scenes. The goal is to offer a more comprehensive view of the model’s robustness, generalization ability, and interpretability.

\section*{B. Implementation Details}
Our ProCap is implemented in PyTorch and trained for a single epoch using mixed-precision training.
We optimize the model with the AdamW optimizer, setting the weight decay to 0.05 and the momentum parameters $\beta_1 = 0.9$ and $\beta_2 = 0.99$.
A cosine learning rate decay strategy is employed with an initial learning rate of $1 \times 10^{-4}$.
The training process includes 5{,}000 linear warm-up steps, during which the learning rate is gradually increased from $1 \times 10^{-6}$ to the initial learning rate. More detailed training information is presented in~\cref{tab:training_time_and_gpu}.

\section*{C. Extended Results}
\cref{tab:results_all_60_scenes} and~\cref{tab:results_all_5_scenes} extend~\cref{tab:results_60_seen_scenes} and~\cref{tab:results_5_unseen_scenes} in main paper and report detailed comparisons across COCO~\cite{lin2014coco}, nocaps~\cite{Agrawal2019nocaps}, and WHOOPS!~\cite{Bitton-Guetta2023WHOOPS} subsets, which themselves consist of numerous scenes. To complement the evaluation in normal scenes, we further captioned projected content directly on them instead of in physical scenes, as illustrated in~\cref{tab:results_projection_caption}. These results collectively confirm that ProCap provides both improved accuracy and generalization in complex SAR scene caption tasks. Several consistent findings emerge:  
\begin{itemize}
    \item In the scene captioning task across 60 seen scenes, ProCap variants significantly outperform off-the-shelf baselines and demonstrate exceptional robustness in distinguishing physical scenes from projection content.
    
    \item In the projection captioning task across 60 seen scenes, Qwen3-VL-8B-Instruct~{$_{\text{RGBP}}$} demonstrates a decisive advantage over all baselines, particularly in achieving significantly higher CIDEr scores on nocaps and COCO. These results underscore the necessity of our dataset.  
    \item In the unseen scenes, ProCap shows limited gains in the scene captioning task. This is likely due to limited scale of training data or inherent SAR complexities. However, fine-tuning on the RGBP dataset enables the model to significantly outperform baselines in the projection captioning task. This highlights the RGBP dataset's critical role in enhancing performance within complex SAR scenes despite limited scene variety.

\end{itemize}

\begin{figure*}[h]
    \centering
    \includegraphics[width=\textwidth]{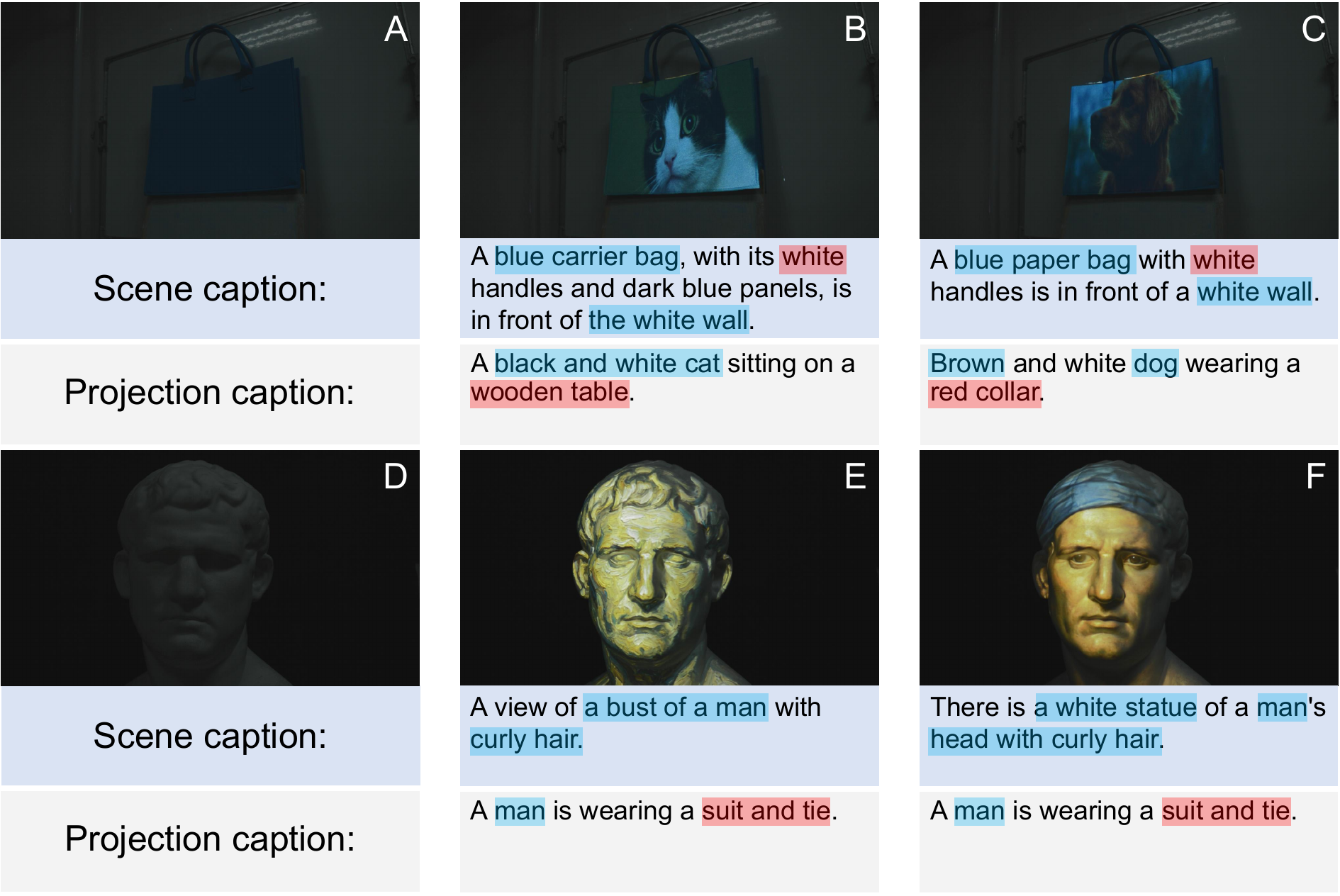}
    \caption{Examples of projection mapping based on TOSHIBA TDP-T100C DLP projector (1024 $\times$ 768) and Nikon D3200 DSLR camera (1280 $\times$ 720) using ProCap~{$_{\text{Vicuna-1.5-7B}}$}. We highlight incorrect captioned objects in \false{red} and correct ones in \true{blue}.
}
    \label{fig:infer_examples}
\end{figure*}

Moreover, for physical scenes where projected content perfectly aligns with the surface of the projected object in SAR applications, we further evaluated ProCap$_{\text{Vicuna-1.5-7B}}$'s understanding of this SAR application. The specific results of these two sets of SAR scenes are shown in~\cref{fig:infer_examples}. This demonstrates the potential future application value of the ProCap framework in the field of projection mapping.

Furthermore, we evaluated the additional 5 unseen scenes using Qwen3-VL-8B-Instruct~{$_{\text{RGBP}}$}, with both the vision encoder and language model adapted via LoRA. As shown in~\cref{tab:results_addtional_5_scenes}, the performance on the new ProCams setups is nearly identical to that achieved on the original 5 evaluation sets in~\cref{tab:results_all_5_scenes}.
The supplementary experiments reinforce several key insights: 
\begin{itemize}
\item The segmentation-first strategy is essential for disentangled scene understanding in SAR contexts. 

\item ProCap maintains robust performance under cross-domain shifts, adversarial content, and unseen scenarios. 

\item The observed performance margins confirm the broader applicability of ProCap as a specialized framework for SAR scenes.
\end{itemize}
Overall, these additional results validate that the proposed segmentation-based, region-aware framework is not only effective but also necessary for advancing visual language understanding in spatial augmented reality.  

\begin{table}[h]
\centering
\caption{Methods on training time and used GPUs.}
\label{tab:training_time_and_gpu}
\setlength{\tabcolsep}{6pt}
\begin{tabular}{lcc}
\toprule
\textbf{Method} & \textbf{Training time} & \textbf{GPUs} \\
\midrule

ProCap~{$_{\text{TinyLlama-1.1B}}$~\cite{zhang2024tinyllama}} & $\sim$4.5h & 3 RTX4060Ti \\
ProCap~{$_{\text{OpenLLaMA-3B}}$~\cite{openlm2023openllama}} & $\sim$4.5h & 3 RTX4060Ti \\
ProCap~{$_{\text{OPT-2.7B}}$~\cite{zhang2022optopenpretrainedtransformer}} & $\sim$4.3h & 3 RTX4060Ti \\
ProCap~{$_{\text{Llama-3.3-8B}}$~\cite{llama3modelcard}} & $\sim$4.2h & 1 A100 \\
ProCap~{$_{\text{Vicuna-1.5-7B}}$~\cite{zheng2023judgingllmasajudgemtbenchchatbot}}& $\sim$4.2h & 1 A100 \\
RGBP~{$_{\text{Qwen3-VL-8B-Instruct}}$~\cite{bai2025qwen3vltechnicalreport}} & $\sim$6.0h & 1 RTX PRO 6000 Blackwell \\

\bottomrule

\end{tabular}
\end{table}

\begin{table*}[t]
    \centering
    \renewcommand{\arraystretch}{1.2} 
    \caption{Performance comparison on the \textbf{60 seen scenes} with unseen projected content in main paper. Metrics are reported as BLEU@4 (B@4), METEOR (M), CIDEr (C), SPICE (S). Results are averaged on the 60 scenes, and the best results are in \textbf{bold}.}
    \label{tab:results_all_60_scenes}

\resizebox{\linewidth}{!}
{
\begin{tabular}{cl rrrr rrrrrrrr rr}
\toprule[1pt]
\multirow{3}{*}{\textbf{Task}} & \multirow{3}{*}{\textbf{Method}}
& \multicolumn{4}{c}{\textbf{COCO}} & \multicolumn{8}{c}{\textbf{nocaps val}} & \multicolumn{2}{c}{\textbf{WHOOPS!}} \\
& & \multicolumn{4}{c}{Test} & \multicolumn{2}{c}{In-domain} & \multicolumn{2}{c}{Near-domain} & \multicolumn{2}{c}{Out-domain} & \multicolumn{2}{c}{Overall} & \multicolumn{2}{c}{Test} \\
\cmidrule(lr){3-6} \cmidrule(lr){7-14} \cmidrule(lr){15-16}
& & \multicolumn{1}{c}{B@4 $\uparrow$} & \multicolumn{1}{c}{M $\uparrow$} & \multicolumn{1}{c}{C $\uparrow$} & \multicolumn{1}{c}{S $\uparrow$} & \multicolumn{1}{c}{C $\uparrow$} & \multicolumn{1}{c}{S $\uparrow$} & \multicolumn{1}{c}{C $\uparrow$} & \multicolumn{1}{c}{S $\uparrow$} & \multicolumn{1}{c}{C $\uparrow$} & \multicolumn{1}{c}{S $\uparrow$} & \multicolumn{1}{c}{C $\uparrow$} & \multicolumn{1}{c}{S $\uparrow$} & \multicolumn{1}{c}{C $\uparrow$} & \multicolumn{1}{c}{S $\uparrow$} \\
\midrule
\multirow{5}{*}[-8em]{%
  \centering
  \rotatebox[origin=c]{90}{%
  \shortstack[c]{\footnotesize Scene captioning}
  }%
}

& FastVLM-0.5B~\cite{fastvlm2025} & 4.76 & 14.25 & 1.72 & 9.20 & 1.87 & 8.83 & 1.78 & 8.93 & 2.16 & 10.65 & 1.60 & 9.47 & 1.59 & 10.17  \\
& FastVLM-1.5B~\cite{fastvlm2025}  & 4.75 & 14.15 & 1.92 & 8.77 & 1.98 & 8.44 & 1.96 & 8.64 & 2.33 & 10.42 & 1.73 & 9.16 & 1.74 & 9.92 \\
& FastVLM-7B~\cite{fastvlm2025}  & 5.62 & 15.63 & 2.31 & 10.31 & 2.51 & 10.13 & 2.55 & 10.16 & 2.87 & 12.08 & 2.21 & 10.79 & 2.21 & 11.98 \\
& Qwen2.5-VL-3B-Instruct~\cite{bai2025qwen25vltechnicalreport} & 7.95 & 18.56 & 2.20 & 11.99 & 2.28 & 11.93 & 2.29 & 11.86 & 2.58 & 12.85 & 2.00 & 12.21 & 1.99 & 13.04 \\
& Qwen2.5-VL-7B-Instruct~\cite{bai2025qwen25vltechnicalreport} & 8.34 & 19.15 & 2.02 & 12.45 & 2.23 & 12.80 & 2.24 & 12.70 & 2.25 & 13.03 & 1.85 & 12.85 & 1.87 & 13.67 \\
& Qwen3-VL-2B-Instruct~\cite{bai2025qwen3vltechnicalreport} & 8.58 & 19.18 & 2.34 & 12.17 & 2.42 & 11.79 & 2.48 & 11.82 & 2.72 & 12.89 & 2.12 & 12.17 & 2.01 & 12.83 \\
& Qwen3-VL-4B-Instruct~\cite{bai2025qwen3vltechnicalreport} & 5.99 & 18.32 & 2.38 & 11.68 & 2.66 & 11.80 & 2.64 & 11.95 & 2.64 & 12.07 & 2.19 & 11.94 & 2.02 & 12.15 \\
& Qwen3-VL-8B-Instruct~\cite{bai2025qwen3vltechnicalreport} & 7.19 & 19.75 & 2.38 & 13.03 & 2.50 & 13.20 & 2.50 & 12.99 & 2.60 & 13.24 & 2.09 & 13.15 & 1.97 & 13.07 \\
& InternVL3-1B~\cite{zhu2025internvl3exploringadvancedtraining}  & 7.57 & 17.81 & 1.97 & 11.64 & 1.93 & 11.11 & 1.92 & 11.00 & 2.13 & 12.48 & 1.67 & 11.53 & 1.67 & 12.16 \\
& InternVL3-2B~\cite{zhu2025internvl3exploringadvancedtraining}  & 10.39 & 19.24 & 3.07 & 12.37 & 3.15 & 12.11 & 3.11 & 12.08 & 3.76 & 13.41 & 2.80 & 12.53 & 2.74 & 13.04 \\
& InternVL3-8B~\cite{zhu2025internvl3exploringadvancedtraining}  & 10.00 & 19.16 & 3.49 & 13.00 & 3.52 & 13.03 & 3.57 & 12.93 & 3.88 & 13.75 & 3.06 & 13.24 & 3.01 & 13.89 \\

& ProCap~{$_{\text{TinyLlama-1.1B}}$ (ours)} & \textbf{36.50} & \textbf{29.23} & \textbf{70.27} & 21.92 & \textbf{51.50} & 21.54 & 53.99 & 23.38 & \textbf{50.03} & 23.21 & 35.21 & 24.21 & \textbf{69.95} & 21.87 \\
& ProCap~{$_{\text{OpenLLaMA-3B}}$ (ours)} & 35.96 & 28.89 & 69.43 & 22.01 & 49.69 & 20.80 & \textbf{55.63} & 22.64 & 49.30 & 23.17 & 37.14 & 22.75 & 69.46 & 22.08 \\
& ProCap~{$_{\text{Llama-3.3-8B}}$ (ours)} & 33.36 & 28.82 & 35.64 & 22.24 & 35.16 & 22.19 & 35.46 & 22.21 & 35.07 & 22.20 & 35.21 & 22.20 & 35.49 & 22.14 \\
& ProCap~{$_{\text{Vicuna-1.5-7B}}$ (ours)} & 34.15 & 29.13 & 36.17 & 22.75 & 36.98 & 22.94 & 36.22 & 22.84 & 36.07 & 22.85 & 36.39 & 22.88 & 36.53 & 22.70 \\
& ProCap~{$_{\text{OPT-2.7B}}$ (ours)} & 35.95 & 28.75 & 28.01 & \textbf{24.34} & 28.44 & \textbf{24.44} & 28.34 & \textbf{24.46} & 28.46 & \textbf{24.48} & 27.63 & \textbf{24.45} & 28.19 & \textbf{24.11} \\
& Qwen3-VL-8B-Instruct~{$_{\text{RGBP}}$~\cite{bai2025qwen3vltechnicalreport}} & 36.38 & 29.18 & 37.81 & 23.22 & 37.41 & 23.47 & 37.80 & 23.48 & 37.32 & 23.54 & \textbf{37.48} & 23.49 & 37.70 & 23.59 \\
\midrule
\multirow{5}{*}[-8em]{%
  \centering
  \rotatebox[origin=c]{90}{%
  \shortstack[c]{\footnotesize Projection captioning}
  }%
}

& FastVLM-0.5B~\cite{fastvlm2025} & 4.58 & 14.18 & 7.17 & 7.43 & 6.91 & 6.39 & 6.93 & 6.32 & 3.89 & 4.35 & 6.16 & 5.68 & 5.54 & 5.85 \\
& FastVLM-1.5B~\cite{fastvlm2025} & 4.75 & 14.02 & 7.21 & 7.31 & 6.82 & 6.79 & 6.20 & 6.36 & 3.64 & 4.48 & 5.72 & 5.88 & 4.14 & 6.09 \\
& FastVLM-7B~\cite{fastvlm2025} & 5.26 & 14.15 & 7.65 & 7.33 & 7.11 & 6.77 & 7.30 & 6.88 & 5.91 & 4.69 & 7.01 & 6.12 & 6.25 & 6.77 \\
& Qwen2.5-VL-3B-Instruct~\cite{bai2025qwen25vltechnicalreport} & 6.85 & 17.33 & 8.41 & 10.52 & 10.67 & 9.00 & 12.33 & 9.11 & 9.04 & 7.48 & 10.79 & 8.53 & 7.37 & 8.92 \\
& Qwen2.5-VL-7B-Instruct~\cite{bai2025qwen25vltechnicalreport} & 7.32 & 18.19 & 14.07 & 11.43 & 16.52 & 9.67 & 18.14 & 9.54 & 15.95 & 8.53 & 17.11 & 9.25 & 12.15 & 10.27 \\
& Qwen3-VL-2B-Instruct~\cite{bai2025qwen3vltechnicalreport} & 7.97 & 18.80 & 10.73 & 12.47 & 14.41 & 11.11 & 16.47 & 10.93 & 13.39 & 9.12 & 14.85 & 10.38 & 9.82 & 11.05 \\
& Qwen3-VL-4B-Instruct~\cite{bai2025qwen3vltechnicalreport} & 5.85 & 18.18 & 10.59 & 11.72 & 14.64 & 11.14 & 17.06 & 10.90 & 13.25 & 9.37 & 14.93 & 10.47 & 10.35 & 10.91 \\
& Qwen3-VL-8B-Instruct~\cite{bai2025qwen3vltechnicalreport} & 6.10 & 17.83 & 11.56 & 11.63 & 15.38 & 10.31 & 17.64 & 10.66 & 13.92 & 9.07 & 15.57 & 10.01 & 11.19 & 10.97 \\
& InternVL3-1B~\cite{zhu2025internvl3exploringadvancedtraining} & 6.63 & 17.04 & 8.02 & 10.33 & 10.95 & 8.91 & 12.35 & 8.79 & 8.49 & 6.58 & 10.78 & 8.10 & 6.89 & 9.12 \\
 & InternVL3-2B~\cite{zhu2025internvl3exploringadvancedtraining} & 11.06 & 18.81 & 37.71 & 11.67 & 37.68 & 9.50 & 42.89 & 9.36 & 36.14 & 6.93 & 39.66 & 8.60 & 34.43 & 11.02 \\
 & InternVL3-8B~\cite{zhu2025internvl3exploringadvancedtraining} & 11.14 & 20.25 & 38.68 & 13.67 & 36.94 & 10.99 & 42.05 & 11.12 & 37.89 & 8.75 & 39.31 & 10.29 & 34.55 & 12.27 \\

& ProCap~{$_{\text{TinyLlama-1.1B}}$ (ours)} & 15.68 & 16.61 & 54.37 & 9.75 & 39.62 & 5.74 & 29.77 & 4.79 & 17.69 & 3.64 & 30.45 & 4.80 & 11.88 & 3.62 \\
& ProCap~{$_{\text{OpenLLaMA-3B}}$ (ours)} & 24.18 & 20.59 & 76.98 & 13.58 & 57.61 & 8.43 & 42.73 & 6.82 & 24.10 & 4.57 & 42.06 & 6.66 & 19.37 & 5.43 \\
& ProCap~{$_{\text{Llama-3.3-8B}}$ (ours)} & 25.60 & 21.18 & 79.10 & 14.10 & 52.57 & 8.18 & 37.16 & 6.47 & 24.17 & 4.89 & 38.57 & 6.51 & 20.09 & 5.71 \\
& ProCap~{$_{\text{Vicuna-1.5-7B}}$ (ours)} & 24.58 & 21.05 & 78.99 & 13.83 & 55.19 & 8.32 & 39.75 & 6.68 & 22.85 & 4.75 & 39.93 & 6.59 & 22.33 & 6.02 \\
& ProCap~{$_{\text{OPT-2.7B}}$ (ours)} & 18.05 & 16.83 & 57.72 & 10.21 & 46.48 & 6.91 & 30.40 & 5.41 & 13.03 & 3.19 & 30.58 & 5.17 & 15.04 & 4.40 \\
& Qwen3-VL-8B-Instruct~{$_{\text{RGBP}}$~\cite{bai2025qwen3vltechnicalreport}} & \textbf{36.37} & \textbf{27.75} & \textbf{127.58} & \textbf{21.14} & \textbf{99.52} & \textbf{13.17} & \textbf{107.20} & \textbf{13.74} & \textbf{98.00} & \textbf{12.68} & \textbf{102.67} & \textbf{13.19} & \textbf{80.46} & \textbf{16.07} \\

\bottomrule[1pt]
\end{tabular}
}
\end{table*}

\begin{table*}[t]
    \centering
    \renewcommand{\arraystretch}{1.2} 
    \caption{Performance comparison on the \textbf{5 unseen scenes} with unseen projected content in main paper. Metrics are reported as BLEU@4 (B@4), METEOR (M), CIDEr (C), SPICE (S). Results are averaged on the 5 scenes, and the best results are in \textbf{bold}.}
    \label{tab:results_all_5_scenes}

\resizebox{\linewidth}{!}
{
\begin{tabular}{cl rrrr rrrrrrrr rr}
\toprule[1pt]
\multirow{3}{*}{\textbf{Task}} & \multirow{3}{*}{\textbf{Method}}
& \multicolumn{4}{c}{\textbf{COCO}} & \multicolumn{8}{c}{\textbf{nocaps val}} & \multicolumn{2}{c}{\textbf{WHOOPS!}} \\
& & \multicolumn{4}{c}{Test} & \multicolumn{2}{c}{In-domain} & \multicolumn{2}{c}{Near-domain} & \multicolumn{2}{c}{Out-domain} & \multicolumn{2}{c}{Overall} & \multicolumn{2}{c}{Test} \\
\cmidrule(lr){3-6} \cmidrule(lr){7-14} \cmidrule(lr){15-16}
& & \multicolumn{1}{c}{B@4 $\uparrow$} & \multicolumn{1}{c}{M $\uparrow$} & \multicolumn{1}{c}{C $\uparrow$} & \multicolumn{1}{c}{S $\uparrow$} & \multicolumn{1}{c}{C $\uparrow$} & \multicolumn{1}{c}{S $\uparrow$} & \multicolumn{1}{c}{C $\uparrow$} & \multicolumn{1}{c}{S $\uparrow$} & \multicolumn{1}{c}{C $\uparrow$} & \multicolumn{1}{c}{S $\uparrow$} & \multicolumn{1}{c}{C $\uparrow$} & \multicolumn{1}{c}{S $\uparrow$} & \multicolumn{1}{c}{C $\uparrow$} & \multicolumn{1}{c}{S $\uparrow$} \\
\midrule
\multirow{5}{*}[-8em]{%
  \centering
  \rotatebox[origin=c]{90}{%
  \shortstack[c]{\footnotesize Scene captioning}
  }%
}

& FastVLM-0.5B~\cite{fastvlm2025} & 2.16 & 10.98 & 1.44 & 5.42 & 1.42 & 5.20 & 1.50 & 4.96 & 1.90 & 6.62 & 1.32 & 5.58 & 1.02 & 6.78 \\
& FastVLM-1.5B~\cite{fastvlm2025} & 1.88 & 10.90 & 0.98 & 5.36 & 1.14 & 5.14 & 1.14 & 5.98 & 1.48 & 8.16 & 1.04 & 6.44 & 0.84 & 8.50 \\
& FastVLM-7B~\cite{fastvlm2025} & 2.10 & 11.64 & 1.24 & 6.00 & 1.60 & 6.30 & 1.48 & 6.02 & 1.88 & 7.80 & 1.38 & 6.72 & 1.34 & 8.66 \\
& Qwen2.5-VL-3B-Instruct~\cite{bai2025qwen25vltechnicalreport} & 4.40 & 14.08 & 2.14 & 7.50 & 2.36 & 7.50 & 2.46 & 7.30 & 2.52 & 8.58 & 2.10 & 7.78 & 1.82 & 8.94 \\
& Qwen2.5-VL-7B-Instruct~\cite{bai2025qwen25vltechnicalreport} & 5.50 & 14.88 & 2.12 & 8.38 & 2.32 & 8.32 & 2.36 & 8.46 & 2.60 & 9.26 & 2.04 & 8.68 & 1.70 & 8.96 \\
& Qwen3-VL-2B-Instruct~\cite{bai2025qwen3vltechnicalreport} & 4.48 & 15.42 & 2.36 & 9.32 & 2.56 & 8.66 & 2.66 & 8.64 & 2.66 & 9.24 & 2.18 & 8.86 & 1.86 & 8.94 \\
& Qwen3-VL-4B-Instruct~\cite{bai2025qwen3vltechnicalreport} & 5.06 & 16.60 & 3.44 & 9.08 & 4.02 & 8.64 & 3.98 & 8.72 & 4.12 & 9.26 & 3.42 & 8.88 & 2.98 & 9.04 \\
& Qwen3-VL-8B-Instruct~\cite{bai2025qwen3vltechnicalreport} & 5.32 & \textbf{17.32} & 3.68 & \textbf{12.26} & 4.12 & \textbf{11.88} & 3.96 & \textbf{11.86} & 4.16 & \textbf{12.32} & 3.52 & \textbf{12.00} & 3.08 & \textbf{12.46} \\
& InternVL3-1B~\cite{zhu2025internvl3exploringadvancedtraining}  & 4.34 & 14.74 & 1.26 & 8.20 & 1.28 & 7.76 & 1.28 & 7.56 & 1.76 & 9.68 & 1.20 & 8.34 & 1.06 & 8.80 \\
& InternVL3-2B~\cite{zhu2025internvl3exploringadvancedtraining}  & \textbf{5.90} & 15.18 & 2.22 & 9.24 & 2.54 & 8.84 & 2.58 & 8.98 & 3.62 & 10.82 & 2.40 & 9.54 & 2.12 & 10.40 \\
& InternVL3-8B~\cite{zhu2025internvl3exploringadvancedtraining}  & 4.84 & 15.82 & 2.38 & 10.14 & 2.70 & 9.94 & 2.80 & 9.86 & 3.02 & 10.72 & 2.34 & 10.16 & 2.24 & 10.98 \\

& ProCap~{$_{\text{TinyLlama-1.1B}}$ (ours)} & 2.48 & 11.38 & 5.02 & 5.53 & 4.94 & 8.34 & 4.74 & 8.34 & \textbf{6.08} & 8.50 & 4.06 & 8.34 & \textbf{7.48} & 5.15 \\
& ProCap~{$_{\text{OpenLLaMA-3B}}$ (ours)} & 1.50 & 11.73 & 3.78 & 7.77 & 4.88 & 7.16 & 4.96 & 7.28 & 5.56 & 7.14 & 4.02 & 7.10 & 5.73 & 6.60 \\
& ProCap~{$_{\text{Llama-3.3-8B}}$ (ours)} & 2.22 & 12.10 & 2.92 & 7.44 & 3.20 & 6.94 & 3.02 & 6.68 & 3.32 & 7.12 & 2.64 & 6.92 & 2.46 & 6.90 \\
& ProCap~{$_{\text{Vicuna-1.5-7B}}$ (ours)} & 1.90 & 11.66 & 2.94 & 7.50 & 2.80 & 7.46 & 2.76 & 7.18 & 3.04 & 7.26 & 2.34 & 7.30 & 2.22 & 7.12 \\
& ProCap~{$_{\text{OPT-2.7B}}$ (ours)} & 4.10 & 14.46 & \textbf{5.80} & 9.44 & \textbf{5.78} & 9.06 & 5.62 & 9.24 & 5.50 & 9.06 & 4.82 & 9.12 & 4.68 & 8.90 \\
& Qwen3-VL-8B-Instruct~{$_{\text{RGBP}}$~\cite{bai2025qwen3vltechnicalreport}} & 5.64 & 13.64 & 4.82 & 9.40 & 5.76 & 8.84 & \textbf{6.04} & 8.96 & 5.92 & 8.80 & \textbf{4.96} & 8.86 & 4.00 & 9.42 \\

\midrule
\multirow{5}{*}[-8em]{%
  \centering
  \rotatebox[origin=c]{90}{%
  \shortstack[c]{\footnotesize Projection captioning}
  }%
}

& FastVLM-0.5B~\cite{fastvlm2025} & 6.28 & 16.28 & 11.54 & 9.38 & 8.40 & 7.60 & 10.48 & 7.92 & 6.42 & 5.66 & 8.94 & 7.02 & 8.72 & 8.06 \\
& FastVLM-1.5B~\cite{fastvlm2025} & 6.20 & 15.38 & 10.00 & 8.92 & 7.58 & 7.84 & 7.48 & 7.86 & 4.60 & 6.02 & 6.84 & 7.26 & 4.88 & 7.70 \\
& FastVLM-7B~\cite{fastvlm2025} & 6.44 & 16.36 & 9.70 & 9.38 & 8.62 & 8.50 & 9.70 & 9.20 & 8.80 & 6.62 & 9.44 & 8.14 & 7.84 & 9.08 \\
& Qwen2.5-VL-3B-Instruct~\cite{bai2025qwen25vltechnicalreport} & 9.18 & 20.52 & 11.88 & 14.24 & 14.52 & 11.64 & 16.80 & 11.48 & 12.18 & 10.30 & 14.66 & 11.14 & 9.52 & 11.78 \\
& Qwen2.5-VL-7B-Instruct~\cite{bai2025qwen25vltechnicalreport} & 9.32 & 21.16 & 19.52 & 14.60 & 20.56 & 11.64 & 24.60 & 11.96 & 23.12 & 11.34 & 23.00 & 11.62 & 16.28 & 12.70 \\
& Qwen3-VL-2B-Instruct~\cite{bai2025qwen3vltechnicalreport} & 10.40 & 22.18 & 14.10 & 16.14 & 17.96 & 13.34 & 21.50 & 13.24 & 19.16 & 12.04 & 19.60 & 12.84 & 13.04 & 13.78 \\
& Qwen3-VL-4B-Instruct~\cite{bai2025qwen3vltechnicalreport} & 6.82 & 20.44 & 12.86 & 14.32 & 17.46 & 12.90 & 20.34 & 13.16 & 17.74 & 11.72 & 18.46 & 12.58 & 12.72 & 12.88 \\
& Qwen3-VL-8B-Instruct~\cite{bai2025qwen3vltechnicalreport} & 7.52 & 20.58 & 14.34 & 14.68 & 18.42 & 12.30 & 21.46 & 13.04 & 18.06 & 11.64 & 19.20 & 12.34 & 13.90 & 13.10 \\
& InternVL3-1B~\cite{zhu2025internvl3exploringadvancedtraining} & 8.88 & 19.72 & 10.56 & 13.64 & 14.02 & 11.00 & 15.86 & 10.88 & 11.86 & 8.58 & 14.12 & 10.16 & 9.28 & 11.92 \\
& InternVL3-2B~\cite{zhu2025internvl3exploringadvancedtraining} & 13.78 & 21.22 & 46.94 & 14.24 & 46.24 & 11.24 & 52.88 & 11.00 & 46.54 & 8.26 & 49.14 & 10.18 & 41.40 & 13.34 \\
& InternVL3-8B~\cite{zhu2025internvl3exploringadvancedtraining} & 13.10 & 22.90 & 45.84 & 16.80 & 42.98 & 12.82 & 52.10 & 13.38 & 46.80 & 11.00 & 47.74 & 12.42 & 43.10 & 14.82 \\

& ProCap~{$_{\text{TinyLlama-1.1B}}$ (ours)} & 13.13 & 17.92 & 46.90 & 12.60 & 42.32 & 6.10 & 33.64 & 5.26 & 19.70 & 3.96 & 32.36 & 5.10 & 15.23 & 4.57 \\
& ProCap~{$_{\text{OpenLLaMA-3B}}$ (ours)} & 19.90 & 21.22 & 66.03 & 14.90 & 58.22 & 8.66 & 48.00 & 7.34 & 27.10 & 5.20 & 44.72 & 7.06 & 24.65 & 6.80 \\
& ProCap~{$_{\text{Llama-3.3-8B}}$ (ours)} & 27.82 & 21.98 & 85.02 & 15.02 & 56.68 & 8.56 & 38.20 & 6.62 & 27.02 & 5.42 & 41.12 & 6.88 & 21.00 & 5.92 \\
& ProCap~{$_{\text{Vicuna-1.5-7B}}$ (ours)} & 26.20 & 22.08 & 86.26 & 14.88 & 58.70 & 8.84 & 43.46 & 7.18 & 27.32 & 5.30 & 43.94 & 7.10 & 24.44 & 6.60 \\
& ProCap~{$_{\text{OPT-2.7B}}$ (ours)} & 19.70 & 17.78 & 63.40 & 10.96 & 51.70 & 7.48 & 32.22 & 5.78 & 16.50 & 3.82 & 34.18 & 5.68 & 16.72 & 4.82 \\
& Qwen3-VL-8B-Instruct~{$_{\text{RGBP}}$~\cite{bai2025qwen3vltechnicalreport}} & \textbf{38.18} & \textbf{28.90} & \textbf{136.60} & \textbf{22.24} & \textbf{104.50} & \textbf{13.54} & \textbf{113.76} & \textbf{14.36} & \textbf{103.94} & \textbf{13.30} & \textbf{108.66} & \textbf{13.76} & \textbf{85.82} & \textbf{17.02} \\

\bottomrule[1pt]
\end{tabular}
}
\end{table*}

\begin{table*}[t]
    \centering
    \renewcommand{\arraystretch}{1.2} 
    \caption{Performance comparison on unseen projection captioning task on the 60 seen scenes (ProCap~{$_{\text{Vicuna-1.5-7B, seen scene}}$}) and the 5 unseen scenes (ProCap~{$_{\text{Vicuna-1.5-7B, unseen scene}}$}). In particular, we captioned projected content themselves (ProCap~{$_{\text{Vicuna-1.5-7B, projected content}}$}) instead of in physical scenes. Metrics are reported as BLEU@4 (B@4), METEOR (M), CIDEr (C), SPICE (S). Results are averaged, and the best results are in \textbf{bold}.}
    \label{tab:results_projection_caption}

\resizebox{\linewidth}{!}{
\begin{tabular}{l rrrr rrrrrrrr rr}
\toprule[1pt]
\multirow{3}{*}{\textbf{Projection Content}}
& \multicolumn{4}{c}{\textbf{COCO}} & \multicolumn{8}{c}{\textbf{nocaps val}} & \multicolumn{2}{c}{\textbf{WHOOPS!}} \\
& \multicolumn{4}{c}{Test} & \multicolumn{2}{c}{In-domain} & \multicolumn{2}{c}{Near-domain} & \multicolumn{2}{c}{Out-domain} & \multicolumn{2}{c}{Overall} & \multicolumn{2}{c}{Test} \\
\cmidrule(lr){2-5} \cmidrule(lr){6-13} \cmidrule(lr){14-15}
& \multicolumn{1}{c}{B@4 $\uparrow$} & \multicolumn{1}{c}{M $\uparrow$} & \multicolumn{1}{c}{C $\uparrow$} & \multicolumn{1}{c}{S $\uparrow$} & \multicolumn{1}{c}{C $\uparrow$} & \multicolumn{1}{c}{S $\uparrow$} & \multicolumn{1}{c}{C $\uparrow$} & \multicolumn{1}{c}{S $\uparrow$} & \multicolumn{1}{c}{C $\uparrow$} & \multicolumn{1}{c}{S $\uparrow$} & \multicolumn{1}{c}{C $\uparrow$} & \multicolumn{1}{c}{S $\uparrow$} & \multicolumn{1}{c}{C $\uparrow$} & \multicolumn{1}{c}{S $\uparrow$} \\
\midrule
ProCap~{$_{\text{Vicuna-1.5-7B, seen scene}}$} & 24.58 & 21.05 & 78.99 & 13.83 & 55.19 & 8.32 & 39.75 & 6.68 & 22.85 & 4.75 & 39.93 & 6.59 & 22.33 & 6.02 \\
ProCap~{$_{\text{Vicuna-1.5-7B, unseen scene}}$} & \textbf{26.20} & 22.08 & 86.26 & 14.88 & 58.70 & 8.84 & 43.46 & 7.18 & \textbf{27.32} & \textbf{5.30} & \textbf{43.94} & \textbf{7.10} & 24.44 & 6.60 \\
ProCap~{$_{\text{Vicuna-1.5-7B, projected content}}$} & \textbf{26.20} & \textbf{23.00} & \textbf{89.80} & \textbf{15.60} & \textbf{63.6} & \textbf{9.40} & \textbf{45.60} & \textbf{7.90} & 18.70 & 3.90 & 43.50 & \textbf{7.10} & \textbf{27.60} & \textbf{7.40} \\

\bottomrule[1pt]
\end{tabular}
}
\end{table*}

\begin{table}[!tb]
    \centering
    \small
    \renewcommand{\arraystretch}{1} 
    \caption{Performance comparison on the additional \textbf{5 unseen scenes} with unseen projected content, using an EPSON EB-C2050WN 3LCD projector (1600 × 1200) and a Nikon D3200 DSLR camera (1080 × 720). Metrics are reported as CIDEr (C), SPICE (S). Results are averaged on the 5 scenes, and the best results are in \textbf{bold}. ``Proj." denotes Projection.}
    \label{tab:results_addtional_5_scenes}
{
\begin{tabular}{cl rr rr rr}
\toprule[1pt]
\multirow{3}{*}{\textbf{Task}} & \multirow{3}{*}{\textbf{Method}}
& \multicolumn{2}{c}{\textbf{COCO}} & \multicolumn{2}{c}{\textbf{nocaps val}} & \multicolumn{2}{c}{\textbf{WHOOPS!}} \\
& & \multicolumn{2}{c}{Test} & \multicolumn{2}{c}{Overall} & \multicolumn{2}{c}{Test} \\
\cmidrule(lr){3-4} \cmidrule(lr){5-6} \cmidrule(lr){7-8}
& & \multicolumn{1}{c}{C $\uparrow$} & \multicolumn{1}{c}{S $\uparrow$} & \multicolumn{1}{c}{C $\uparrow$} & \multicolumn{1}{c}{S $\uparrow$} & \multicolumn{1}{c}{C $\uparrow$} & \multicolumn{1}{c}{S $\uparrow$} \\
\midrule
\multirow{2}{*}[0.46em]{%
  \centering
  \rotatebox[origin=c]{90}{%
  \shortstack[c]{\footnotesize Scene}
  }%
}
& Qwen3-VL-8B-Instruct~{$_{\text{w/~~ RGBP fine-tuned}}$} & \textbf{3.18} & 6.84 & \textbf{2.96} & 6.80 & \textbf{2.92} & 7.46 \\
& Qwen3-VL-8B-Instruct~{$_{\text{w/o RGBP fine-tuned}}$} & 0.48 & \textbf{9.84} & 0.38 & \textbf{9.58} & 0.32 & \textbf{10.20} \\
\midrule
\multirow{2}{*}{%
  \centering
  \rotatebox[origin=c]{90}{%
  \shortstack[c]{\footnotesize Proj.}
  }%
}
& Qwen3-VL-8B-Instruct~{$_{\text{w/~~ RGBP fine-tuned}}$} & \textbf{133.50} & \textbf{22.18} & \textbf{106.38} & \textbf{13.64} & \textbf{87.10} & \textbf{17.34} \\
& Qwen3-VL-8B-Instruct~{$_{\text{w/o RGBP fine-tuned}}$} & 4.96 & 15.32 & 8.08 & 13.54 & 5.80 & 13.18 \\

\bottomrule[1pt]
\end{tabular}
}
\end{table}

\end{document}